\documentclass[10pt,twocolumn,letterpaper]{article}

\usepackage[pagenumbers]{wacv} 

\usepackage{graphicx}
\usepackage{amsmath}
\usepackage{amssymb}
\usepackage{booktabs}
\usepackage{mathtools}
\usepackage{array}
\usepackage{soul}
\newcolumntype{P}[1]{>{\centering\arraybackslash}p{#1}}
\usepackage{float}
\usepackage{multirow}
\usepackage{float}      
\usepackage{afterpage}
\usepackage{scalerel,xparse}
\usepackage{capt-of,etoolbox}
\usepackage{tikz}

\usepackage[pagebackref,breaklinks,colorlinks]{hyperref}
\usepackage[capitalize]{cleveref}
\crefname{section}{sec.}{secs.}
\Crefname{section}{Sec.}{Secs.}
\crefname{table}{tab.}{tabs.}
\Crefname{table}{Tab.}{Tabs.}
\crefname{figure}{fig.}{figs.}
\Crefname{figure}{Fig.}{Figs.}
\crefname{equation}{eq.}{eqs.}
\Crefname{equation}{Eq.}{Eqs.}

\def\wacvPaperID{896} 
\def\confName{WACV}
\def\confYear{2025}

\newcommand{\method}{\textsc{ZeroComp}\xspace}

\newcommand{\myparagraph}[1]{\noindent\textbf{#1}}

\begin{document}

\title{\method: Zero-shot Object Compositing from Image Intrinsics via Diffusion}



\author{Zitian Zhang$^1$ \quad
Frédéric Fortier-Chouinard$^1$ \quad
Mathieu Garon$^2$ \\
Anand Bhattad$^3$ \quad
Jean-Fran\c{c}ois Lalonde$^1$ \\
$^1$Université Laval, $^2$Depix Technologies, $^3$Toyota Technological Institute at Chicago \\
\small{\texttt{\url{https://lvsn.github.io/ZeroComp}}}
}

\twocolumn[{%
\renewcommand\twocolumn[1][]{#1}%
\maketitle
\begin{center}
    \centering
    \captionsetup{type=figure}
    \vspace{-5mm}
    \footnotesize
    \setlength{\tabcolsep}{1pt}
    \begin{tabular}{ccc}
        \includegraphics[width=0.25\textwidth]{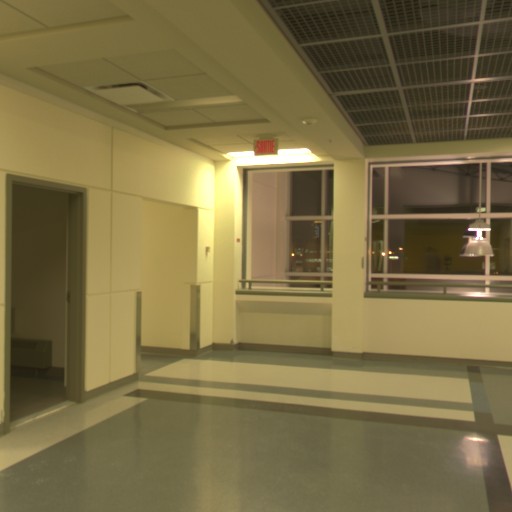} & 
        \includegraphics[width=0.25\textwidth]{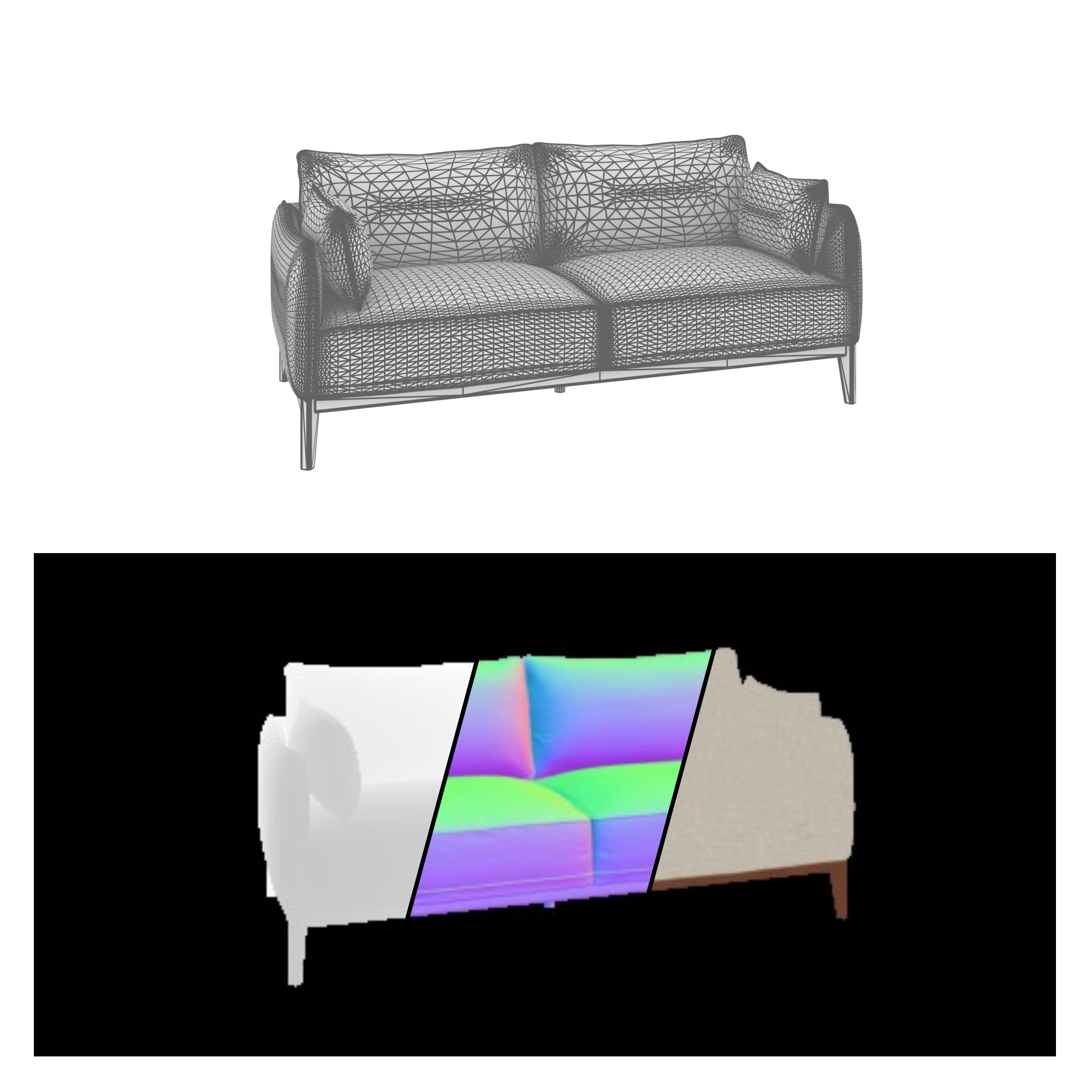} & 
        \includegraphics[width=0.25\textwidth]{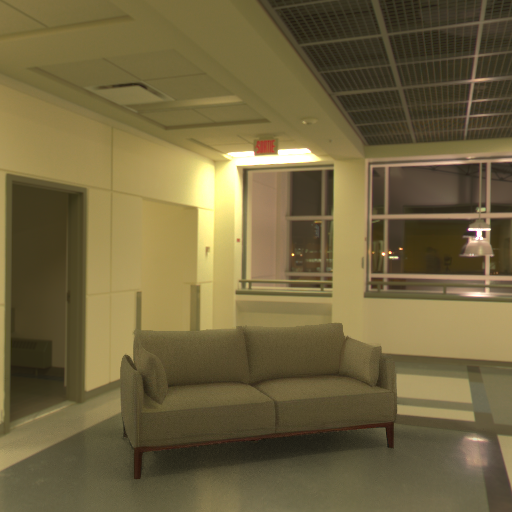}  \\
        (a) Target image &  (b) Intrinsic maps & (c) Predicted composite \\
    \end{tabular}
    \vspace{-5pt}
    \captionof{figure}{From (a) a target background image and (b) available intrinsic maps (depth, normals, albedo) rendered from a 3D model, our method \method generates (c) a realistic composite, without access to the scene geometry or lighting, and without being trained specifically for object compositing. \method realistically shades the object and adds a compelling shadow.}
    \label{fig:teaser}
\end{center}%
}]
\begin{abstract}
We present \method, an effective zero-shot 3D object compositing approach that does not require paired composite-scene images during training. Our method leverages ControlNet to condition from intrinsic images and combines it with a Stable Diffusion model to utilize its scene priors, together operating as an effective rendering engine. During training, \method uses intrinsic images based on geometry, albedo, and masked shading, all without the need for paired images of scenes with and without composite objects. Once trained, it seamlessly integrates virtual 3D objects into scenes, adjusting shading to create realistic composites. We develop a high-quality evaluation dataset and demonstrate that \method outperforms methods using explicit lighting estimations and generative techniques in quantitative and human perception benchmarks. Additionally, \method extends to real and outdoor image compositing, even when trained solely on synthetic indoor data, showcasing its effectiveness in image compositing.

\vspace{-15pt}
\end{abstract}

\section{Introduction}
Compositing 3D objects into real photographs has become an essential task in diverse fields such as image editing and visual effects.
To achieve high realism, the virtual object---defined as a 3D asset with geometry, texture, etc.---must interact with the lighting and elements from the real target scene. To this end, Debevec~\cite{debevec1998rendering} introduced ``image-based lighting'' (IBL), a three-step approach for compositing of 3D objects into real images: 1) capture the high dynamic range (HDR) lighting properties of the target scene to use as a virtual light source; 2) approximate the surrounding target scene geometry to catch virtual shadows; and 3) render and composite the virtual object into the scene. Since then, several methods have been proposed to improve upon each of these steps, but the overall approach has remained the same. For example, lighting estimation methods~\cite{gardner2017sigasia} have replaced the need for capturing light probes; depth estimation~\cite{bhat2023zoedepth} and camera auto-calibration~\cite{jin2023perspective} can approximate the target scene geometry. 


Recent works~\cite{chen2023anydoor,zhang2023controlcom} has departed from this three-step procedure and leveraged pre-trained diffusion models such as Stable Diffusion~\cite{rombach2022high} (SD). This creates a simpler, single-pass approach that creates highly realistic images. However, the object is often adjusted, rotated, or even deformed by the SD model, which leads to unpredictable results. 

Addressing the limitations of current IBL- and SD-based compositing methods, we propose \method, which merges the generative power of SD models with the precision of the IBL framework to enable realistic compositing of virtual 3D objects into images. Central to \method is training a ControlNet \cite{zhang2023adding} that leverages a pre-trained SD backbone to render images from an intrinsic decomposition of a given scene. This training results in zero-shot compositing, allowing object insertion into diverse scenes without specific prior training. Consequently, \method ensures realistic compositing of virtual objects, preserving their shape, pose, and texture, without requiring paired  training dataset. This zero-shot capability stems from our rendering-focused training strategy, offering a compositing solution within a single framework.


To achieve this, our key idea is to train \method on a simpler proxy task: given a decomposition of an image into its intrinsic components---depth, normals, albedo, and partial input shading---generate a fully relit image. This network is trained using synthetic datasets like OpenRooms \cite{li2020openrooms} or InteriorVerse \cite{zhu2022learning}. At inference, new 3D objects are inserted into the depth, normals, and albedo layers as naive composites. The trained model then generates a fully-shaded version of the object, faithful to the scene lighting while retaining object identity. In short, \method acts as a neural renderer, specifically trained to generate illumination effects such as shading and cast shadows (\Cref{fig:teaser}).






To rigorously assess our approach, we compile a meticulously curated dataset for evaluating image compositing, utilizing 3D object assets from the Amazon Berkeley Object dataset~\cite{collins2022abo} and background scenes from the Laval HDR Indoor dataset~\cite{gardner2017sigasia}. Extensive evaluation shows that our method competes well with traditional methods relying on explicit lighting estimation and surpasses other SD-based strategies in realism and faithfulness to the object. Given that quantitative metrics often do not correlate well with human perception \cite{giroux2024towards}, we conducted a user study that demonstrated our approach significantly outshines all other methods.
In summary our contributions are: 1) we present \method for zero-shot object compositing, where a ControlNet is trained as a neural renderer, conditioned on intrinsic layers; 2) a novel test dataset for evaluating 3D object compositing methods; 3) a thorough evaluation unifying several methods from the recent literature, including lighting estimation, shadow generation, diffusion models and harmonization approaches; and 4) a user study demonstrating the performance of \method at generating renders more perceptually pleasing than state-of-the-art methods.




\section{Related work}
\label{sec:relwork}

Image compositing has a long history in vision and graphics. Of note, Reinhard et al.~\cite{reinhard2001color} study color-based harmonization, later revisited in \cite{pitie2005n}. Issuing from the pioneering work by Burt and Adelson~\cite{burt1983multiresolution}, seamless object blending has been studied in \cite{perez2003poisson,rother2004grabcut,tao2013error}. Lalonde et al.~\cite{lalonde2007photo} insert compatible objects from a large database into a target scene. Karsch et al.\cite{karsch2011rendering,karsch2014automatic} insert synthetic 3D objects through the estimation of geometry and lighting annotations. Since then, image compositing techniques have approached the problem along different dimensions.

\myparagraph{Using light estimators.} 
This body of work achieves compositing by using HDR lighting for scenes and relighting virtual 3D objects within them. Learning-based methods estimate illumination through both non-parametric models \cite{gardner2017sigasia} and parametric models \cite{gardner2019iccv, hold2017cvpr}. Recent advancements enable editable illumination by merging parametric and non-parametric approaches for lighting adjustments \cite{weber2022editable, li2022physically, wang2022stylelight}.
Efforts to tackle complex scene lighting have led to methods for estimating spatially-varying illumination \cite{garon2019cvpr, li2020cvpr, zhu2022irisformer, srinivasan2020lighthouse, choi2023mair, ge20243d}. Unlike these methods, \method does not rely on explicit lighting estimation and rendering; instead, it jointly learns these tasks.

\myparagraph{Intrinsic images,}
or the process of factorizing an image into albedo and shading, has long been studied~\cite{barrow1978recovering}. More recent approaches also recover other intrinsic maps such as depth and surface normals~\cite{garces2022survey}.
Several approaches try to achieve better decomposition through incorporating human annotations~\cite{zhou2015learning,kovacs2017shading,wu2023measured}, ordinal information~\cite{zoran2015learning,careaga2023ordinal}, physical insights~\cite{yoshida2023light,ramanagopal2024theory}, shade tree structure~\cite{geng2023tree}, or multi-view information~\cite{philip2019multi,ye2023intrinsicnerf}. 
Other methods ~\cite{li2020cvpr,zhu2022irisformer,choi2023mair} estimate shape, spatially-varying lighting, and non-Lambertian surface reflectance for improved compositing. Recent methods have utilized pretrained StyleGAN or SD for the extraction of intrinsic images, either via latent search~\cite{bhattad2023stylegan}, low-rank adaptations~\cite{du2023generative}, through a probabilistic formulation~\cite{kocsis2023iid}, or though ControlNet~\cite{luo2024intrinsicdiffusion}.  Concurrently, RGB$\leftrightarrow$X \cite{zeng2024rgb} decomposes images into intrinsic maps and synthesizes them using material and lighting-aware diffusion models.
Intrinsic images can be used to reshade inserted objects~\cite{bhattad2022cut}. This can be extended by adjusting the shading in the foreground and background separately for compatible composites~\cite{careaga2023intrinsic}. StyLitGAN~\cite{bhattad2024stylitgan, bhattad2023make} relits real images by manipulating StyleGAN latent space. LightIt~\cite{kocsis2024lightit} controls lighting using intrinsics in diffusion models. 
Similarly, we use intrinsic images for compositing and leverage scene priors through pretrained diffusion models.





\myparagraph{Shadows.} Compositing accurate shadows is crucial for photorealism \cite{chuang2003shadow}. Recent advances include shadow generation \cite{liu2020arshadowgan} and harmonization \cite{valencca2023shadow} models. Techniques by Sheng et al. \cite{sheng2022controllable, sheng2021ssn, sheng2023pixht} use pixel height information to generate various light effects, including shadows. In contrast, our approach implicitly learns to generate realistic shadows without shadow-specific supervision, eliminating the need for a separate shadow generation or correction pipeline.

\myparagraph{Image harmonization.} Methods aiming to harmonize the color distribution of inserted objects for coherence with the background include \cite{reinhard2001color, pitie2005n, lalonde2007using, sunkavalli2010multi, tsai2017deep}. Relevant methods also exploit intrinsic images for harmonization \cite{guo2021intrinsic, careaga2023intrinsic}. These approaches typically focus on adjusting object colors and, unlike our work, cannot synthesize shadows.



\begin{figure*}[htpb!]
    \centering
\includegraphics[width=0.95\linewidth]{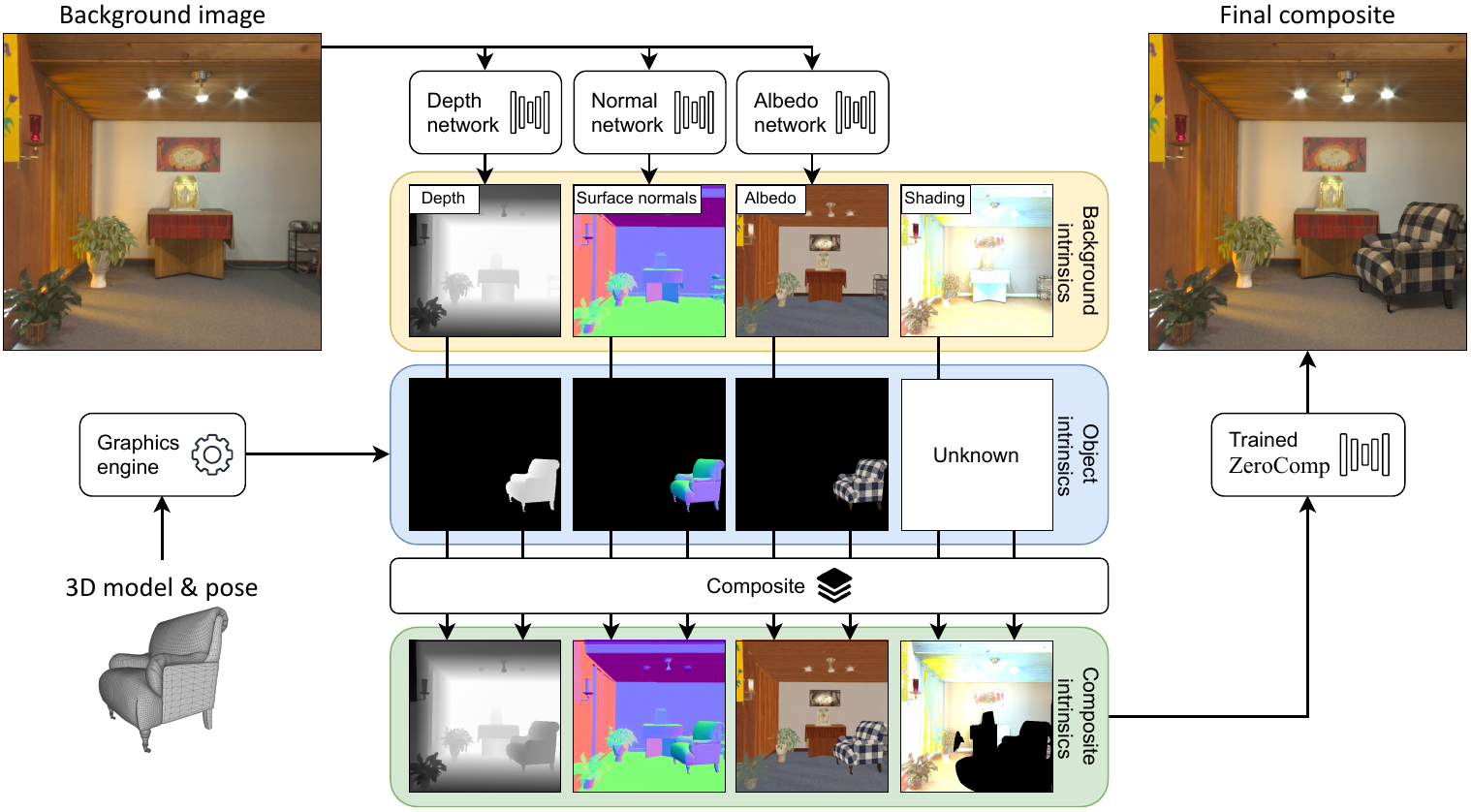}
    \vspace{-5pt}
    \caption{Overview of our zero-shot intrinsic compositing pipeline. The input background image $x_\mathrm{bg}$ (top-left) is first converted to intrinsic layers $\mathbf{i}_\mathrm{bg}$ using specialized networks (top, in yellow). In parallel, the corresponding intrinsic layers of the 3D object $\mathbf{i}_\mathrm{obj}$---except the shading---are rendered using a graphics engine (middle, in blue). Layers are then composited together to obtain the composited intrinsics $\mathbf{i}_\mathrm{comp}$ (bottom, in green). From this, our trained \method renders the final composite $x$ (top-right).}
    \label{fig:pipeline-overview}
    \vspace{-12pt}
\end{figure*}

\myparagraph{Using generative models.} Large generative models like Stable Diffusion (SD)~\cite{rombach2022high} have led to the development of image editing applications by adapting pretrained models through fine-tuning~\cite{hu2021lora}, or by learning adapters~\cite{ye2023ip}. We employ ControlNet~\cite{zhang2023adding} to train a model that accepts intrinsic images and learns to render the final image. 

Recent methods in object compositing have largely leveraged SD. 3DIT \cite{michel2024object} uses language instructions for object insertion but requires paired images for training. CustomNet~\cite{yuan2023customnet} manipulates viewpoint, location, and background, while PhD~\cite{zhang2023paste} employs a two-step harmonization process. ControlCom~\cite{zhang2023controlcom} and ObjectStitch~\cite{song2022objectstitch} focus on embedding manipulation. Paint by Example~\cite{yang2023paint} uses exemplar-guided compositing, and AnyDoor~\cite{chen2023anydoor} uses identity tokens and a frequency-aware feature extractor for detailed object representation. More recent works such as DiffusionLight \cite{phongthawee2024diffusionlight} (lighting estimation), Alchemist \cite{sharma2024alchemist} (material control), DiLightNet \cite{zeng2024dilightnet} (object rendering), and others \cite{zhao2024illuminerf, poirier2024diffusion} (multi-view relighting) rely on specifically-designed training sets. In contrast, our method uniquely integrates intrinsic image handling with ControlNet \cite{zhang2023adding}, enabling zero-shot 3D object compositing without explicit feature manipulation or multi-step processes. 




\section{\method}
\method is a neural renderer that leverages the power of Stable Diffusion~\cite{rombach2022high} and ControlNet~\cite{zhang2023adding}, which is trained to render images from their intrinsic maps, namely depth, normals, albedo and shading. The distinctive aspect of our model is its ability to integrate 3D objects into 2D images without requiring training on paired images of scenes both with and without the objects. We refer to this capability as zero-shot compositing, allowing the model to perform compositing tasks it hasn't been explicitly trained for.

A crucial element of \method involves training it to develop an implicit understanding of the lighting and geometry from the intrinsic maps, enabling it to correct the appearance of the shading where it is missing. During inference, objects can be inserted into scenes using their albedo, normals, and depth maps. For the shading component, regions corresponding to the inserted objects are masked, and our trained \method adjusts this shading to align with the scene's original lighting. An overview of our zero-shot intrinsic compositing approach is illustrated in \Cref{fig:pipeline-overview}.

\subsection{Training \method} 
\label{sec:training}
The training process is meticulously designed to handle the intrinsic components of images where shading is partially available. The model learns to reconstruct scenes from provided intrinsic maps while being conditioned on randomly masked regions within the shading channels. This approach encourages the model to reason about the scene's lighting and geometry autonomously.

\myparagraph{Training data.}
The synthetic OpenRooms~\cite{li2020openrooms} is used as sole training data. This dataset provides depth, normals, albedo, and partial shading information (division between the image and albedo) necessary to understand and recreate complex scenes. Our method can be easily extended to additional intrinsic maps if available, such as roughness and metallic maps in InteriorVerse \cite{zhu2022learning}.

\myparagraph{Shading masks.}
During training, random masks $s$ are generated using a mix of random rectangles/circles (60\% probability) and by removing or keeping the entire shading maps (30\% and 10\% probability resp., see supp.).


\myparagraph{Learning objectives and losses.}
\label{sec:leaning-objectives-and-loss}
The primary training objective is to accurately render images based on intrinsic maps while effectively inpainting missing shading information. This is formalized through a loss function that conditions the rendering process on intrinsic maps and a mask indicating regions for shading inpainting. To improve the fidelity of the hue in the background, we follow \cite{lin2024common} and use zero terminal SNR, v~prediction, DDIM scheduling and trailing timestep selection. The loss function is defined as:
\begin{equation}
    \mathcal{L} = \mathbb{E}_{t,x_0,\epsilon,s} \left\| v_t - \Tilde{v}_t(x_t, \mathbf{i}, s, t) \right\|_2^2 \,,
\end{equation}
where $v_t$ progressively evolves from image to noise over the denoising time steps (see~\cite{lin2024common}). $x_t$ is the generated image at time step~$t$, $\mathbf{i} = \{i_d, i_n, i_a, i_s\}$ is the intrinsic conditions provided to the model, containing depth, normals, albedo and shading resp.,  and $\Tilde{v}_t(\cdot)$ the model prediction. $\mathbb{E}_{t,x_0,\epsilon,s}$ denotes the expectation over: denoising time steps $t$, initial image $x_0$, noise $\epsilon$, and shading mask $s$.


By iteratively minimizing $\mathcal{L}$ across various training steps, \method progressively refines its ability to render realistic images given image intrinsics. We use a pretrained Stable Diffusion 2.1 as the backbone model and condition it on intrinsic inputs $\mathbf{i}$ using ControlNet. The model is trained for 808k steps with a batch size of 32 at a resolution of $512 \times 512$ using 19,709 training samples.

\subsection{Zero-shot object compositing using \method}
\label{sec:zero-shot}

Our goal is to place objects into photographs to achieve a seamless blend without prior access to intrinsic scene information. To accomplish this, we utilize available pretrained, off-the-shelf models that infer the intrinsic properties of the background $\mathbf{i}_\mathrm{bg}$ from the given photograph (\Cref{fig:pipeline-overview}, top). 
Specifically, we employ ZoeDepth~\cite{bhat2023zoedepth} to extract depth maps from the input images. For normals, we leverage StableNormal~\cite{ye2024stablenormal}. Albedo is estimated using Intrinsic Image Diffusion (IID)~\cite{kocsis2024intrinsic}. The shading information is derived by dividing the original image with its albedo. 

For the objects we intend to insert (\Cref{fig:pipeline-overview}, middle), we use the Blender~\cite{blender} graphics engine to render its intrinsic layers $\mathbf{i}_\mathrm{obj}$, except shading which is unknown. As with traditional IBL, this allows the user full control over the object pose and location in the target image. Each intrinsic map from the background and object are then composited together through simple compositing
\begin{equation}
i_{c,\mathrm{comp}} = m \, i_{c,\mathrm{obj}} + (1-m) \, i_{c,\mathrm{bg}} \,,
\label{eqn:comp}
\end{equation}
where $i_c \in \{i_d, i_n, i_a, i_s\}$ denotes one of the intrinsic maps and $m$ the object mask obtained from the graphics engine, resulting in a set of composite intrinsics (\Cref{fig:pipeline-overview}, bottom). Since the depth scale of the object and the background may not match, we align the object footprint depth (planar projection on the vertical axis) with the background depth by fitting an affine transform to the footprint and applying it to  $i_{d,\mathrm{obj}}$. An object also affects surroundings (e.g., by casting shadows), so we mask any pixel from the shading map if the pixel estimated 3D position is within a distance threshold 
\begin{equation}
    d = \lambda ( \max_y m_{y,\mathrm{obj}} - \min_y m_{y,\mathrm{obj}} )\,,
\end{equation}
where $m_{y,\mathrm{obj}}$ represents the (3D) $y$ coordinate of a pixel in the object mask $m$ (obtained from the depth map $i_{d,\mathrm{obj}}$). This threshold is motivated by the fact that the length of a shadow is typically proportional to the object height. In practice, we set the relative shading radius $ \lambda = 1.0$, and explore different values in \Cref{sec:evaluation}. Pixels in the shading map directly above the object are never masked, to avoid unnecessary shadows (on the ceiling, for instance). 

Finally, our trained \method is run on the composite intrinsics to obtain the final output (\Cref{fig:pipeline-overview}, right), where the newly added object appears as a natural part of the original scene. This approach enables the insertion of objects into various scenes with realistic lighting interactions including reshading and casting shadows, achieving zero-shot compositing. For all our experiments, we use seed 469, which was shown in \cite{xu2024goodseedmakesgood} to produce the highest-quality generations in SD among 1000 seeds.

\subsection{Preserving background fidelity}

\begin{figure}[t!]
\centering
\scriptsize
\setlength{\tabcolsep}{1pt}
\begin{tabular}{cccccc}
    \includegraphics[width=0.145\linewidth]{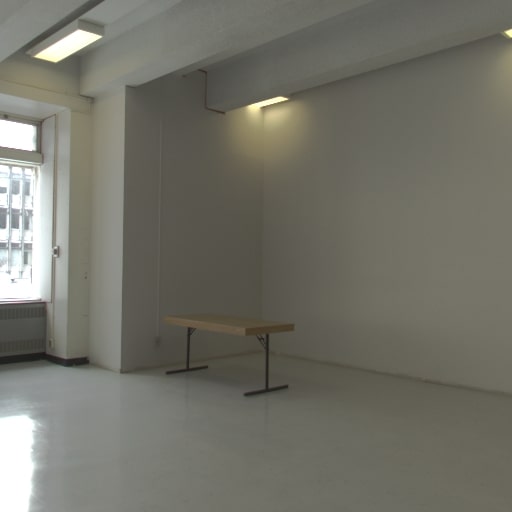}
    \llap{\raisebox{0.8cm}{\frame{\includegraphics[height=1.7cm, trim={0 8cm 15cm 5cm}, clip]{figures/visibility/9C4A8928-2fbd3aa74a_05_crop_B07HZ782D9_vis2/bg.jpg}}}}
    &
    \includegraphics[width=0.145\linewidth]{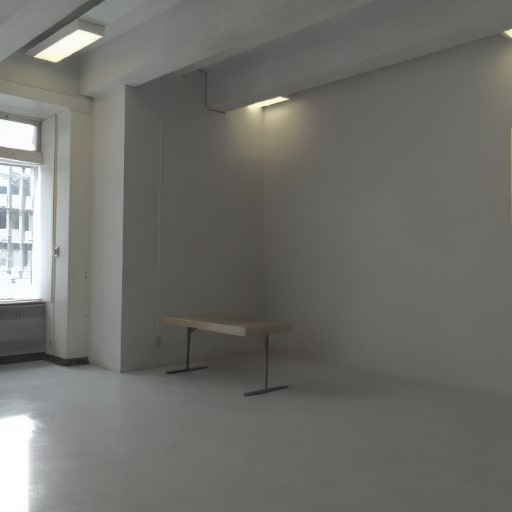}
    \llap{\raisebox{0.8cm}{\frame{\includegraphics[height=1.7cm, trim={0 8cm 15cm 5cm}, clip]{figures/visibility/9C4A8928-2fbd3aa74a_05_crop_B07HZ782D9_vis2/pred_bg.jpg}}}}
    &
    \includegraphics[width=0.145\linewidth]{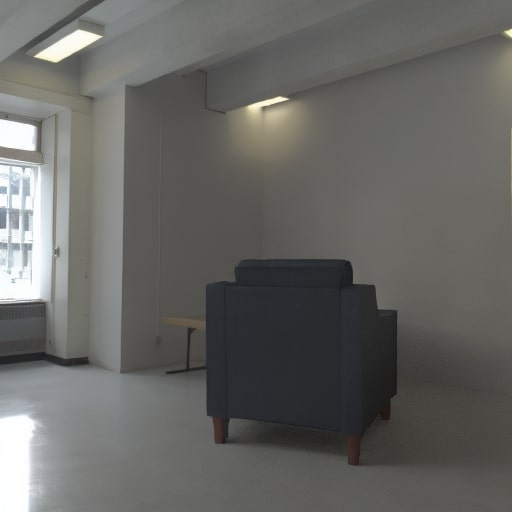}
    \llap{\raisebox{0.8cm}{\frame{\includegraphics[height=1.7cm, trim={0 8cm 15cm 5cm}, clip]{figures/visibility/9C4A8928-2fbd3aa74a_05_crop_B07HZ782D9_vis2/pred.jpg}}}}
    &
    \includegraphics[width=0.145\linewidth]{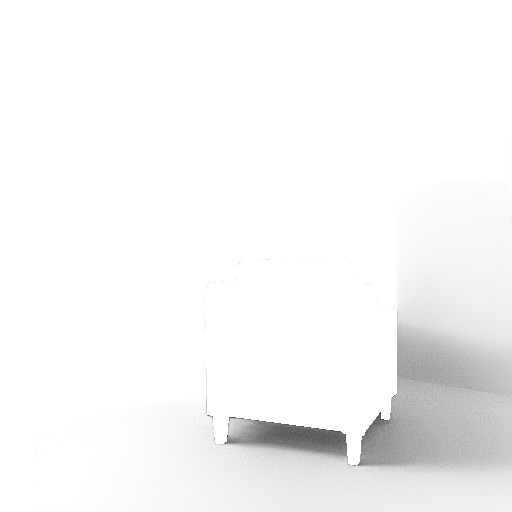} &
    \includegraphics[width=0.145\linewidth]{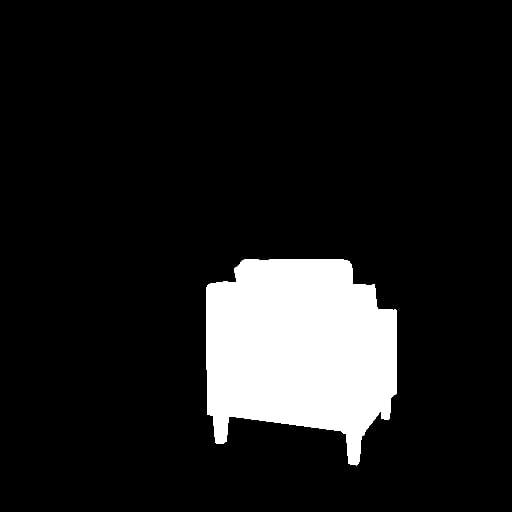} &
    \includegraphics[width=0.145\linewidth]{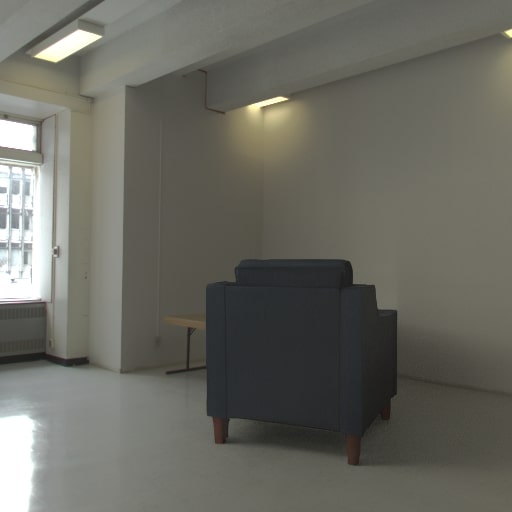}
    \llap{\raisebox{0.8cm}{\frame{\includegraphics[height=1.7cm, trim={0 8cm 15cm 5cm}, clip]{figures/visibility/9C4A8928-2fbd3aa74a_05_crop_B07HZ782D9_vis2/pred_vis.jpg}}}}
    \\
    (a) $x_\mathrm{bg}$ &
    (b) $f_\theta(\mathbf{i}_\mathrm{bg})$ &
    (c) $f_\theta(\mathbf{i}_\mathrm{comp})$ &
    (d) $R$ &
    (e) $m$ & 
    (f) Comp $x$ \\
\end{tabular}
\vspace{-5pt}
\caption{Overview of different components from our full compositing equation in \cref{eqn:comp-final}. For (a) a given target background image $x_\mathrm{bg}$, diffusion models can create artifacts when rendering (b) background $f_\theta(\mathbf{i}_\mathrm{bg})$ and (c) composite $f_\theta(\mathbf{i}_\mathrm{comp})$ intrinsics. To alleviate this, we compute (d) the shadow opacity ratio of predictions $R$ and, together with (e) the object mask $m$, we can create (f) the final artifacts-free composite $x$. Please see the insets (top-right of each column) for a zoomed-in view of the artifacts created.}
\label{fig:background_composition_results}
\vspace{-12pt}
\end{figure}

The ControlNet framework does not guarantee a perfect reconstruction of the background image. As demonstrated in previous studies, small details are susceptible to loss~\cite{zhu2023designing}. To mitigate this, we take inspiration from the differential compositing framework of Debevec~\cite{debevec1998rendering} and generate a shadow opacity ratio from two predictions of \method. Denoting $f_\theta(\mathbf{i})$ as a full inference pass of the model on input intrinsic maps $\mathbf{i} = \{i_d, i_n, i_a, i_s\}$ (containing depth, normals, albedo and shading resp., c.f. \Cref{sec:training}), we compute the shadow opacity ratio $R = 
f_\theta(\mathbf{i}_\mathrm{comp}) / f_\theta(\mathbf{i}_\mathrm{bg}) $,
%
%
where $\mathbf{i}_\mathrm{comp}$ and $\mathbf{i}_\mathrm{bg}$ are the intrinsic maps of the composite and background resp., see \Cref{eqn:comp}. Note that the ratio is computed on grayscale and the result is clamped to $[0,1]$.
We fix the diffusion with the same seed and use the same shading mask on the background $i_s$ to minimize discrepancies between both predictions.
To further avoid unnecessary opacity unrelated to the object, we set the opacity to 1 if they are outside of the shading mask $m$ computed earlier (c.f. \Cref{sec:zero-shot}). A Gaussian blur with kernel $15\times15$ and $\sigma=1.5$ is applied to $m$ to avoid blending artifacts.
The final compositing equation is
\begin{equation}
x = (1-m) \, R \, x_\mathrm{bg} + m \, C \, f(\mathbf{i}_\mathrm{comp}) \,,
\label{eqn:comp-final}
\end{equation}
where $C$ is a color balance factor computed as the average color ratio of background $x_\mathrm{bg}$ and the network output $f(\mathbf{i}_\mathrm{comp})$ to account for global color shifts. In \Cref{fig:background_composition_results}, we compare our composition and direct output of the network. 

\section{Test dataset for 3D object compositing}
\label{sec:dataset}


Evaluating the quality of 3D object composites can be cumbersome, and performing a uniform evaluation of various methods such as lighting estimation, harmonization, or generative techniques is challenging. We require scenes where 1) the background is a real image with known HDR lighting, 2) the scene geometry is defined, 3) a virtual object is correctly positioned, and 4) a realistic rendering exists. Current evaluation datasets often lack some of these requirements: \cite{chen2023anydoor} lack object geometry, \cite{garon2019cvpr} does not include objects, \cite{roberts2021hypersim, li2020openrooms} use synthetic imagery, and 3D Copy-Paste \cite{ge20243d} lacks ground truth HDR lighting. Inspired by \cite{ge20243d}, we propose a simple method for automatically generating a dataset for evaluating 3D object compositing approaches.


Specifically, we leverage the test dataset provided by \cite{weber2022editable, dastjerdi2023everlight}, which contains 2,240 images of $50^\circ$ field of view, extracted from HDR environment maps in the Laval Indoor HDR Dataset~\cite{gardner2017sigasia}, and the result from several lighting estimation methods. For each image, we first find a suitable location to insert a virtual object by computing normals using DSINE \cite{bae2024dsine} on each crop. We detect a support region by selecting normals with an angle less than $15^\circ$ with the up vector. Images with support regions too small (defined as not fitting a circle of 75 pixels radius) are discarded, resulting in 228 admissible background crops.
Next, a 3D object is chosen at random from the ABO dataset~\cite{collins2022abo} and randomly rotated about its vertical axis. The object is scaled to fit its footprint in the support region, ensuring its bounding box is entirely within the camera frustum. Four random objects are rendered for each image, and those with inconsistent geometry, semantics, or unrealistic albedo are discarded, resulting in a total of 213 high-quality images.

\begin{figure*}[!ht]
    \centering
    \footnotesize
    \setlength{\tabcolsep}{0.25pt}
\renewcommand{\arraystretch}{0.25}
\begin{tabular}{ccccccc} 
\includegraphics[width=0.14\textwidth]{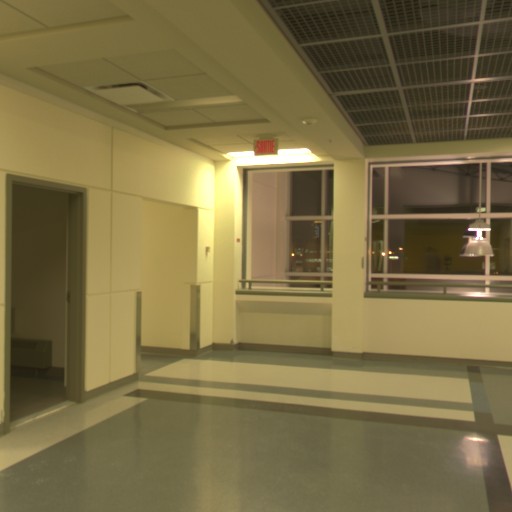} & \includegraphics[width=0.14\textwidth]{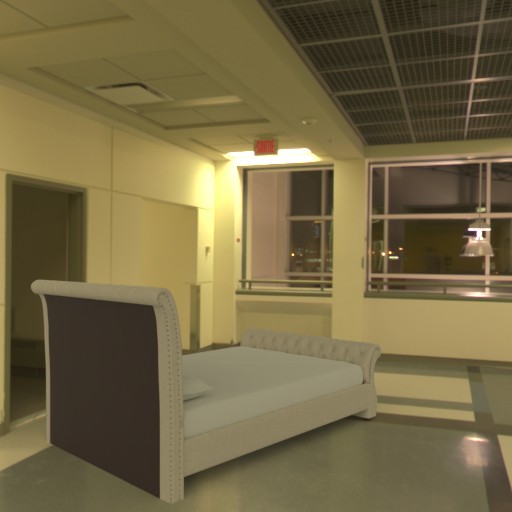} & \includegraphics[width=0.14\textwidth]{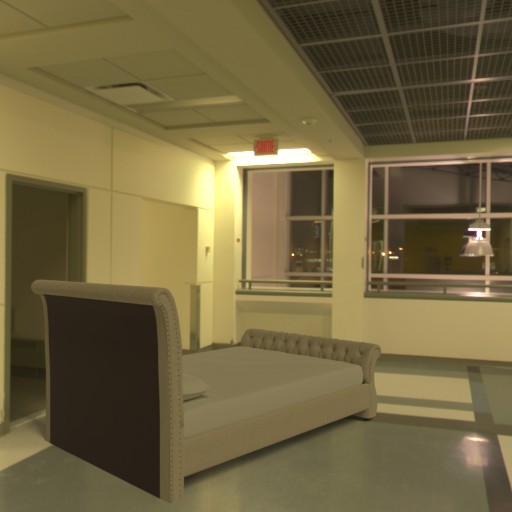} & \includegraphics[width=0.14\textwidth]{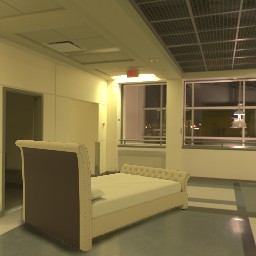} & \includegraphics[width=0.14\textwidth]{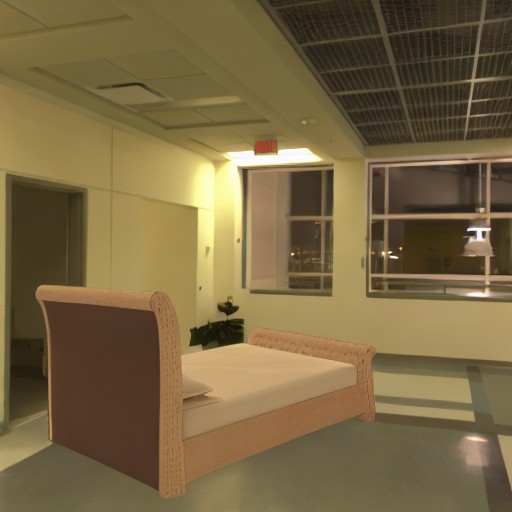} & \includegraphics[width=0.14\textwidth]{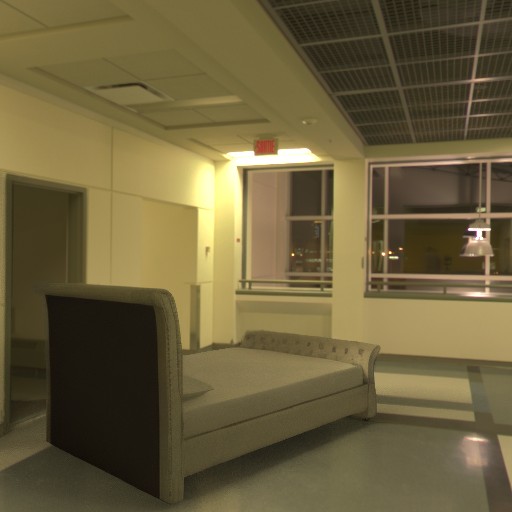} & \includegraphics[width=0.14\textwidth]{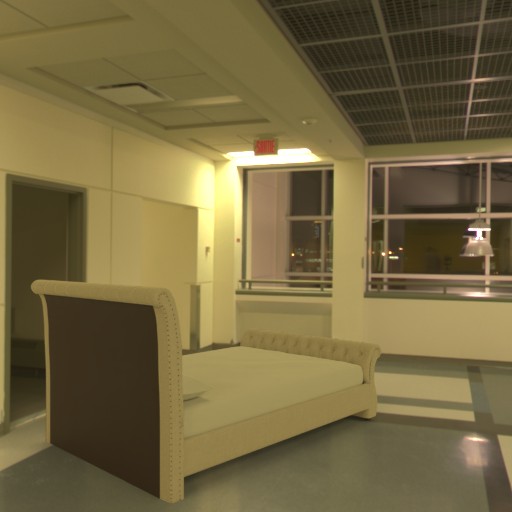}\\  

\includegraphics[width=0.14\textwidth]{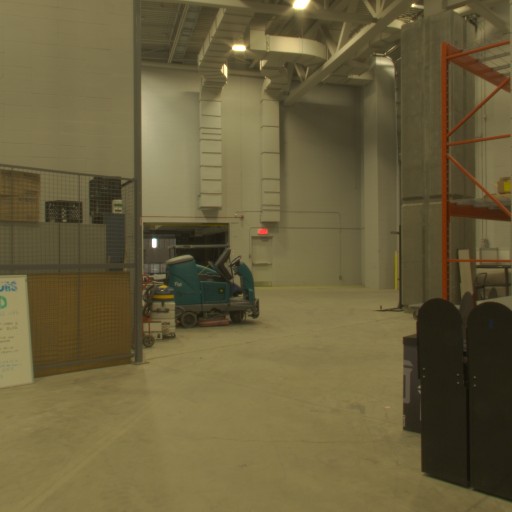} & \includegraphics[width=0.14\textwidth]{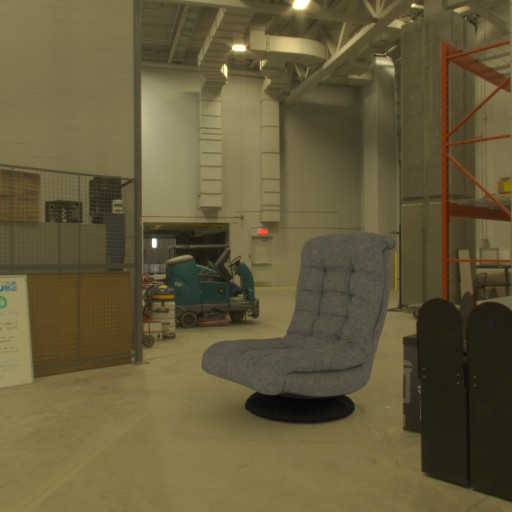} & \includegraphics[width=0.14\textwidth]{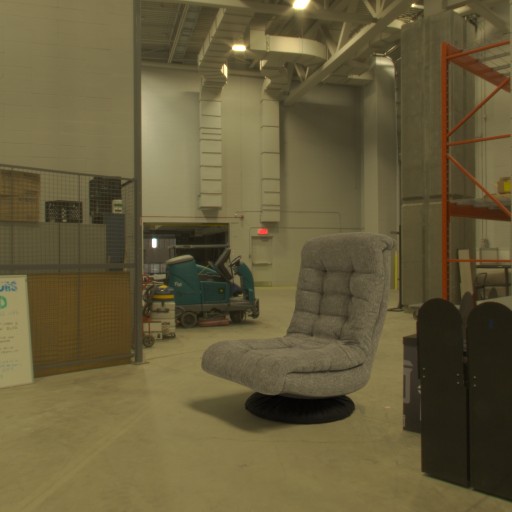} & \includegraphics[width=0.14\textwidth]{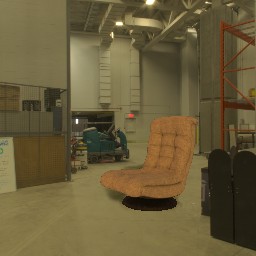} & \includegraphics[width=0.14\textwidth]{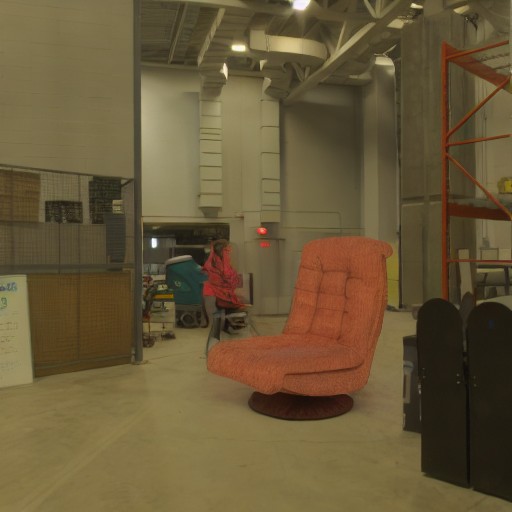} & \includegraphics[width=0.14\textwidth]{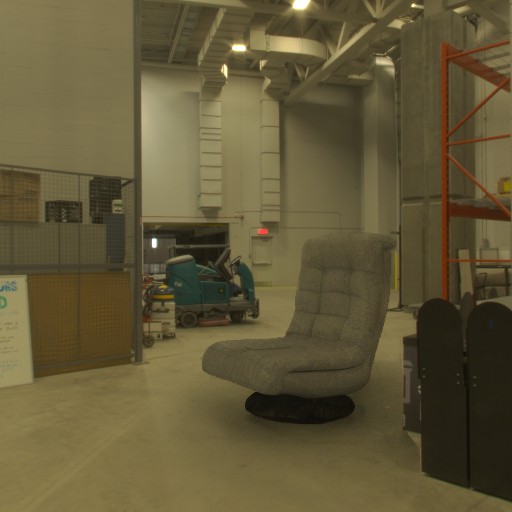} & \includegraphics[width=0.14\textwidth]{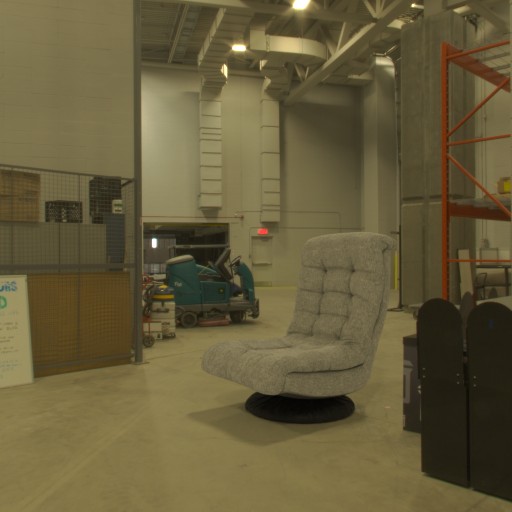}\\  

\includegraphics[width=0.14\textwidth]{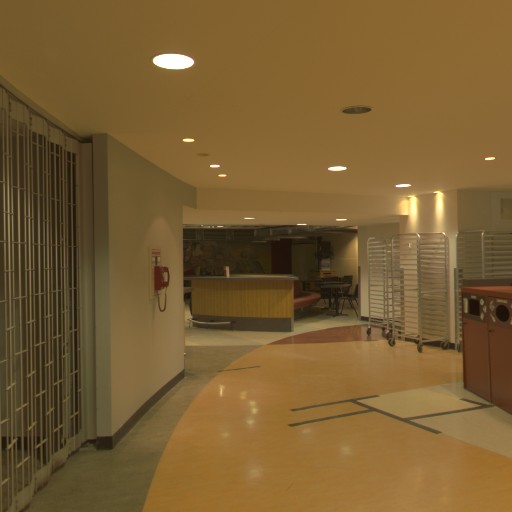} & \includegraphics[width=0.14\textwidth]{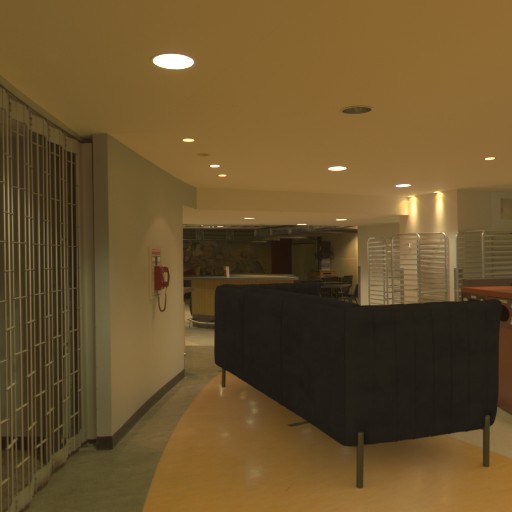} & \includegraphics[width=0.14\textwidth]{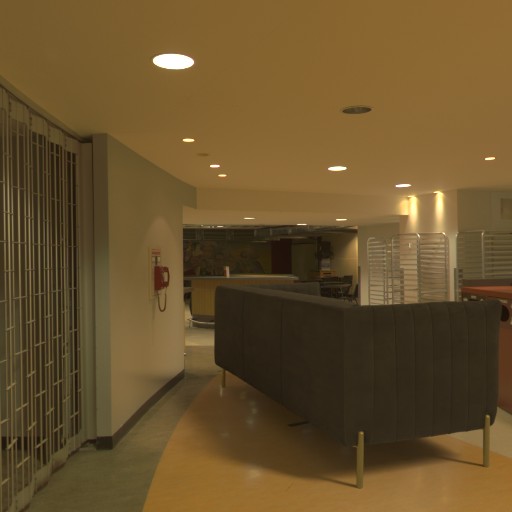} & \includegraphics[width=0.14\textwidth]{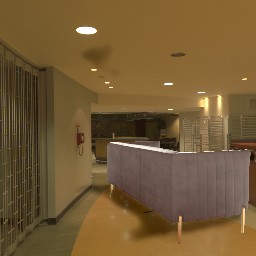} & \includegraphics[width=0.14\textwidth]{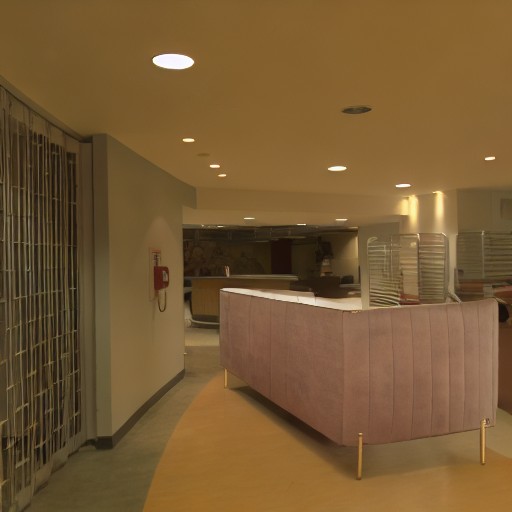} & \includegraphics[width=0.14\textwidth]{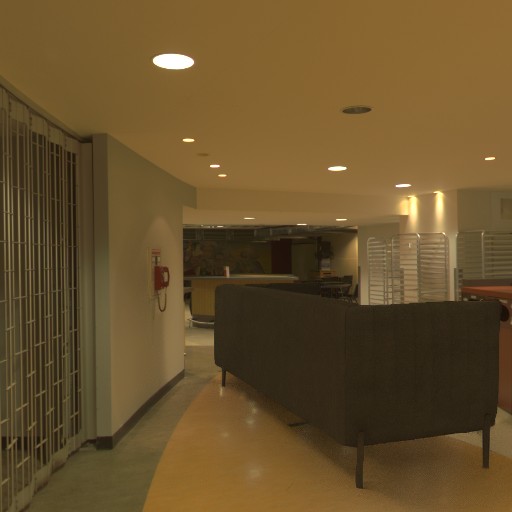} & \includegraphics[width=0.14\textwidth]{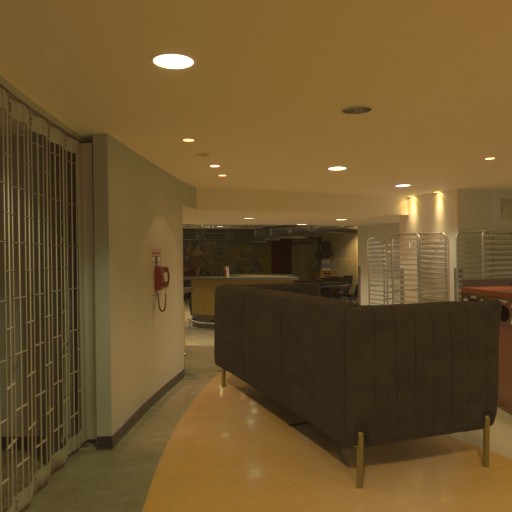}\\  
\includegraphics[width=0.14\textwidth]{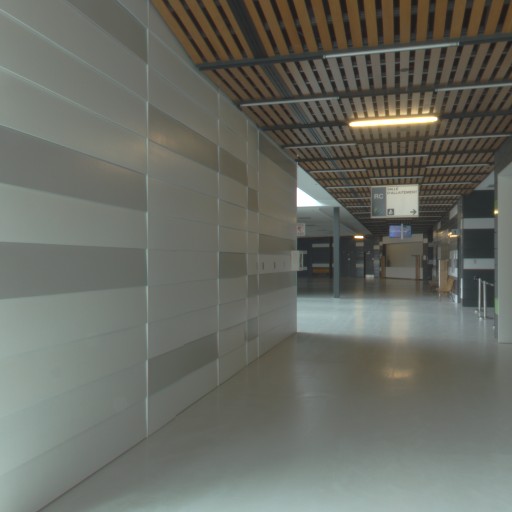} & \includegraphics[width=0.14\textwidth]{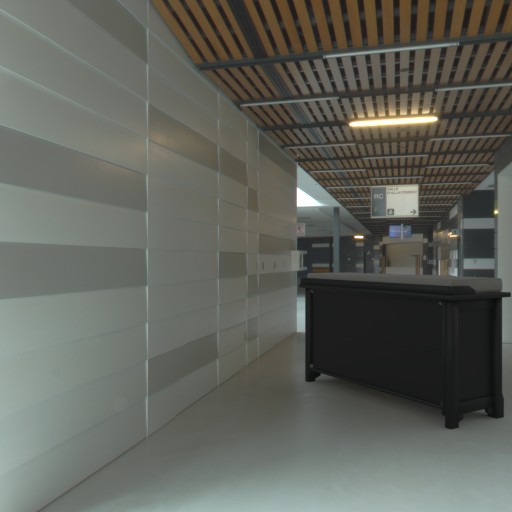} & \includegraphics[width=0.14\textwidth]{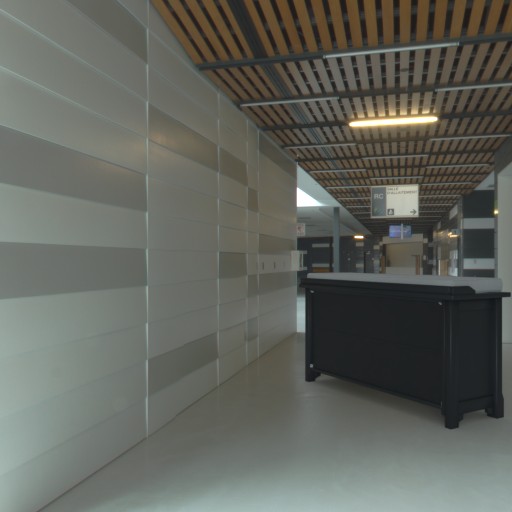} & \includegraphics[width=0.14\textwidth]{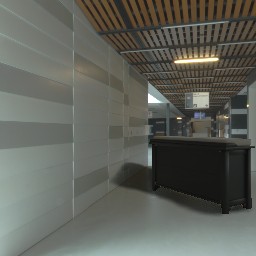} & \includegraphics[width=0.14\textwidth]{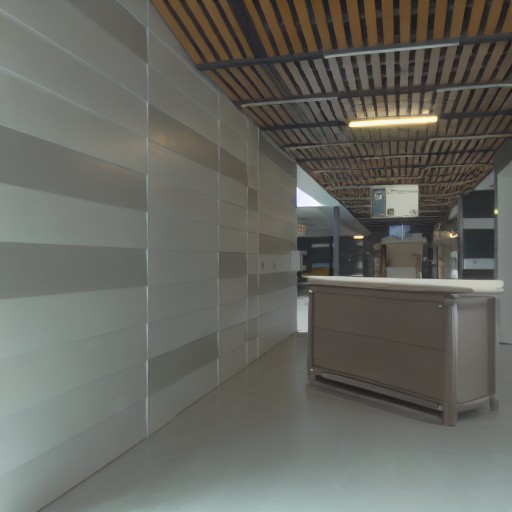} & \includegraphics[width=0.14\textwidth]{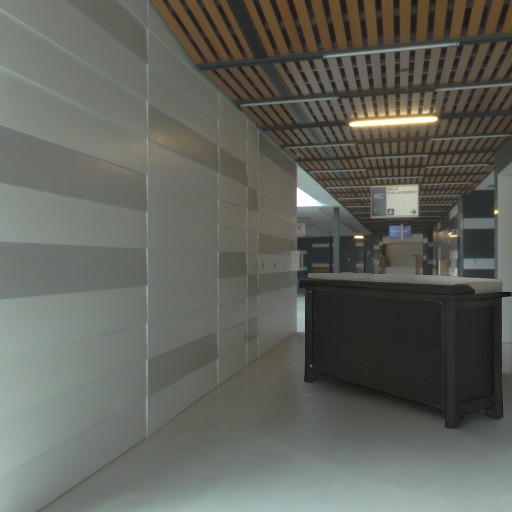} & \includegraphics[width=0.14\textwidth]{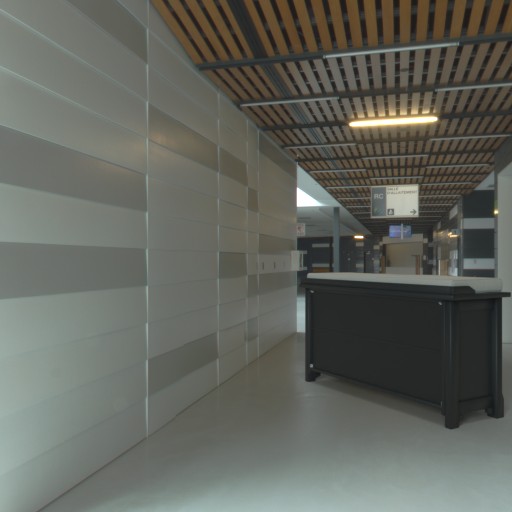}\\  
\includegraphics[width=0.14\textwidth]{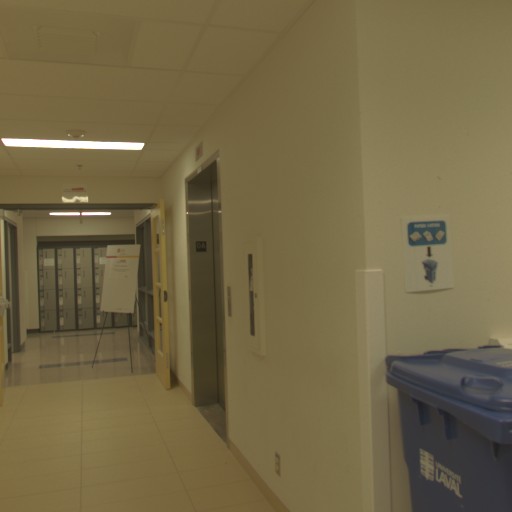} & \includegraphics[width=0.14\textwidth]{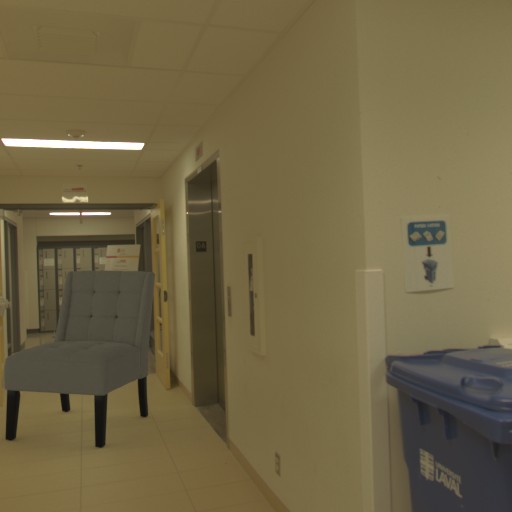} & \includegraphics[width=0.14\textwidth]{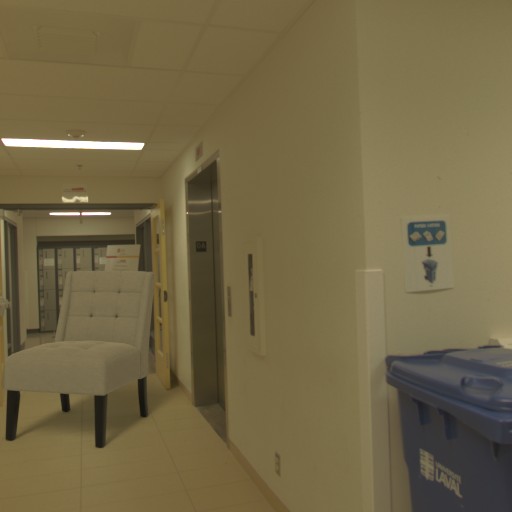} & \includegraphics[width=0.14\textwidth]{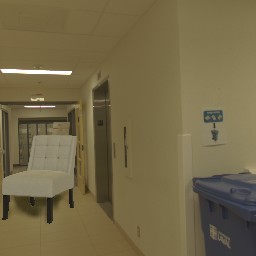} & \includegraphics[width=0.14\textwidth]{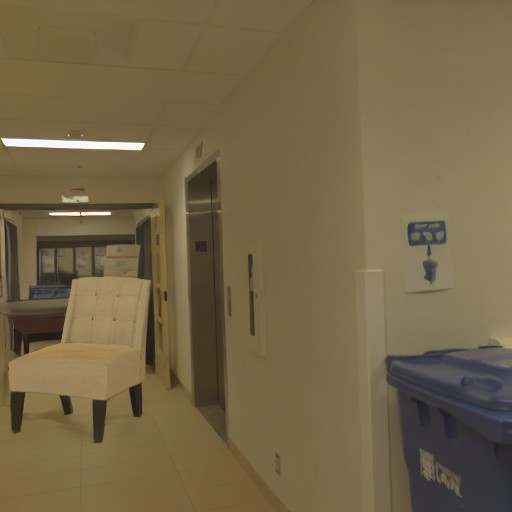} & \includegraphics[width=0.14\textwidth]{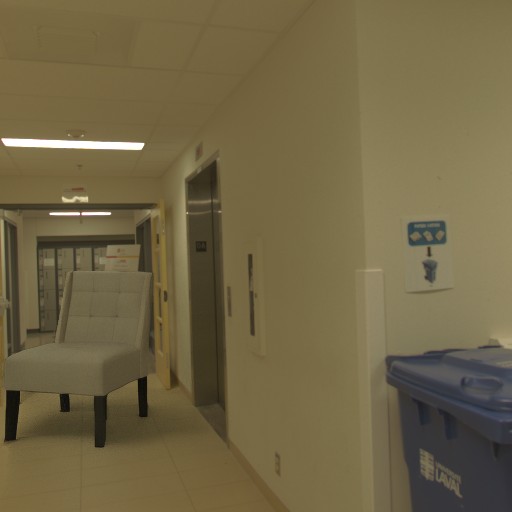} & \includegraphics[width=0.14\textwidth]{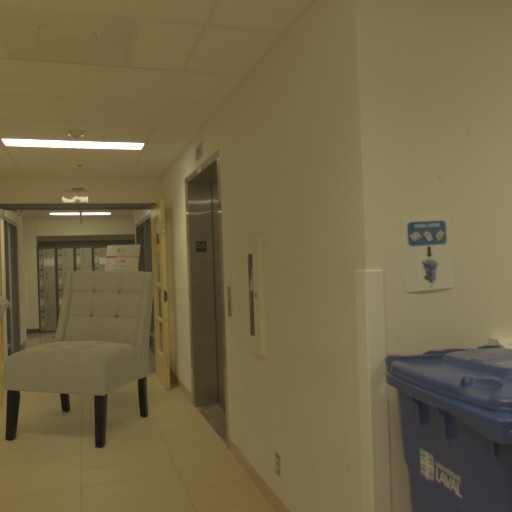}\\  

Background & 
Garon'19~\cite{garon2019cvpr} & EMLight~\cite{zhan2021emlight} & ARShadowGAN~\cite{liu2020arshadowgan} & ControlCom~\cite{zhang2023controlcom} & 
Ours & 
Simulated GT
\end{tabular}
\vspace{-7pt}
\caption{Qualitative comparison with lighting estimation and image-based methods. Results are sorted from worst (top) to best (bottom) PSNR for ``Ours''. Please zoom in and refer to the \textbf{supplementary material} for additional images and methods.}
\label{fig:qualitative_comparison_lighting_methods}
\vspace{-12pt}
\end{figure*}

Finally, we render the object using physically based rendering in Blender \cite{blender}. To account for spatially-varying indoor lighting, we warp the panorama by converting it to a 3D mesh according to its depth map from \cite{rey2022360monodepth}, and reproject it at the center of the object's bounding box. This approach provides high-quality simulated ground truth, enabling the generation of many more scenes than the 20 available in \cite{garon2019cvpr}. The last column of \Cref{fig:qualitative_comparison_lighting_methods} shows a representative subset of realistic image composites. This dataset will be released publicly upon publication of the paper.

\section{Evaluation}
\label{sec:evaluation}

In this section, we conduct a comprehensive evaluation of \method's performance as a neural renderer for zero-shot compositing. Leveraging the evaluation dataset introduced in \Cref{sec:dataset}, we quantitatively and qualitatively compare against state-of-the-art methods on a range of metrics to offer a multi-faceted assessment of image quality. We use standard metrics to compare composites with the simulated ground truth, including Peak Signal-to-Noise Ratio (PSNR), Root Mean Square Error (RMSE) and its scale-invariant version (si-RMSE), Mean Absolute Difference (MAE), Structural Similarity Index Measure (SSIM) \cite{wang2004ssim}, Learned Perceptual Image Patch Similarity (LPIPS) \cite{zhang2018lpips}, and FLIP~\cite{andersson2020flip}. While we agree that perceptual metrics are better suited for our task, several researchers~\cite{goodfellow2014explaining,kettunen2019lpips,ghazanfarilipsim} highlight the vulnerability of neural network-based metrics like LPIPS to noise and adversarial attacks. To mitigate the influence of the rendering noise present in the training datasets~\cite{li2020openrooms,zhu2022learning}, we resize both the test images and references to $256 \times 256$ for all methods on LPIPS. Additionally, FLIP addresses this issue by applying a spatial filter removing high frequency details imperceptible to humans. During evaluation, we demonstrate that our approach achieves performance comparable to most lighting estimation methods, all without explicitly modeling lighting conditions. We also contrast our method with diffusion and intrinsic image-based baselines \cite{chen2023anydoor, zhang2023controlcom, careaga2023intrinsic, liu2020arshadowgan}, showcasing superiority on most metrics.

Given that recent research shows that quantitative metrics do not correlate well with human perception \cite{zamir2019vision,giroux2024towards}, we also conduct a human perceptual study, revealing a clear improvement over all other methods. This highlights \method's ability to produce perceptually plausible results. Finally, we showcase its extensions to material editing, outdoor scenes, and real-world 2D images.



\subsection{Lighting estimation method comparison}
\label{subsec:comparison_lighting_methods}

Traditional lighting-based compositing methods set the benchmark by estimating scene lighting for realistic 3D object insertion. These methods use a full 3D object, a delicately curated model for shadow casting, a physically-based rendering engine, and a suitable lighting representation (e.g., parametric lights\cite{gardner2017sigasia,gardner2019iccv,weber2022editable}, spherical functions~\cite{garon2019cvpr,zhan2021emlight, dastjerdi2023everlight}, etc.). For optimal results, everything must be perfectly aligned.
In contrast, \method only requires placing the object in 2D, generating intrinsics using simple shaders (depth, normals, and albedo), and relies on the network understanding to infer missing information. Despite the task is more challenging, \method achieves competitive results, surpassing many explicit lighting-based techniques \cite{gardner2017sigasia, wang2022stylelight, weber2022editable}, as shown in \Cref{tab:quantative_eval}. Qualitative comparisons in \Cref{fig:qualitative_comparison_lighting_methods} show that \method realistically shades these objects while maintaining their appearance, acting as a strong contender to traditional approaches.

\begin{table}[t!]
\centering
\footnotesize
\setlength{\tabcolsep}{1pt}
\resizebox{\columnwidth}{!}{
\begin{tabular}{llccccccc}
\toprule
& Method & PSNR$_\uparrow$ & RMSE$_\downarrow$ & si-RMSE$_\downarrow$ & SSIM$_\uparrow$ & MAE$_\downarrow$ & LPIPS$_\downarrow$ & FLIP$_\downarrow$ \\
\midrule
\multirow{8}*{\rotatebox{90}{\scriptsize Lighting-based}}

& Gardner'17~\cite{gardner2017sigasia} & 25.9 & 0.0677 & 0.0606 & 0.967 & 0.0211 & 0.0331 & 0.0664 \\
& Garon'19~\cite{garon2019cvpr} & 34.2 & 0.0256 & 0.0251 & 0.986 & 0.0089 & 0.0175 & 0.0440 \\
& Gardner'19~\cite{gardner2019iccv} & 32.3 & 0.0293 & 0.0281 & 0.984 & 0.0091 & 0.0211 & 0.0450 \\
& Everlight~\cite{dastjerdi2023everlight} & 33.3 & 0.0290 & 0.0285 & 0.982 & 0.0102 & 0.0184 & 0.0462 \\
& StyleLight~\cite{wang2022stylelight} & 29.3 & 0.0416 & 0.0399 & 0.976 & 0.0139 & 0.0287 & 0.0580 \\
& Weber'22~\cite{weber2022editable} & 29.6 & 0.0403 & 0.0380 & 0.980 & 0.0130 & 0.0239 & 0.0556 \\
& EMLight~\cite{zhan2021emlight} & 32.7 & 0.0301 & 0.0297 & 0.981 & 0.0104 & 0.0218 & 0.0471 \\
\midrule
\multirow{4}*{\rotatebox{90}{\scriptsize Image-based}} 
& AnyDoor~\cite{chen2023anydoor} & 24.5 & 0.0666 & 0.0657 & 0.883 & 0.0265 & 0.0822 & 0.1098 \\
& ControlCom~\cite{zhang2023controlcom} & 25.5 & 0.0566 & 0.0554 & 0.866 & 0.0283 & 0.0711 & 0.1516 \\
& Careaga'23~\cite{careaga2023intrinsic} & 26.6 & 0.0527 & 0.0498 & 0.965 & 0.0192 & 0.0347 & 0.0884 \\
& ARShadowGAN~\cite{liu2020arshadowgan} & 27.4 & 0.0484 & 0.0467 & 0.907 & 0.0213 & 0.0584 & 0.0994 \\
\midrule
& \method OR & 31.7 & 0.0303 & 0.0295 & 0.970 & 0.0109 & 0.0269 & 0.0538 \\
& \method IV & 33.0 & 0.0259 & 0.0254 & 0.973 & 0.0091 & 0.0246 & 0.0474 \\


\bottomrule
    \end{tabular}
    }
    \vspace{-5pt}
    \caption{Quantitative evaluation. All metrics are computed on the whole image. Different sections indicate methods, from top to bottom: lighting estimation, image-based compositing and ours. OR and IV refer to OpenRooms~\cite{li2020openrooms} and InteriorVerse~\cite{zhu2022learning}.
    }
    \label{tab:quantative_eval}
\end{table}

\subsection{Image-based compositing method comparison}
\label{subsec:comparison_generative_models}
Our evaluation extends to methods that employ intrinsic image decomposition and generative modeling for object compositing. Recent methods like AnyDoor \cite{chen2023anydoor} and ControlCom \cite{zhang2023controlcom} similarly employ a generative framework with a Stable Diffusion backbone, whereas Careaga et al. \cite{careaga2023intrinsic} also rely on intrinsic image decomposition for compositing. Finally, we include ARShadowGAN \cite{liu2020arshadowgan} as an example of shadow generation techniques. All of these methods expect as input an image of the object placed in a different scene. We simulate this setup by rendering the object with a randomly sampled environment map from our test set, and feeding it to the methods, letting them do the task of relighting appropriately based on the target background.

The quantitative results in~\Cref{tab:quantative_eval} (middle bracket) show a superior score for \method in all metrics against image-based compositing methods. We identify two main issues with other approaches: they tend to 1) modify the object itself or its pose~\cite{chen2023anydoor,zhang2023controlcom}; or 2) generate shadows of limited quality~\cite{careaga2023intrinsic,liu2020arshadowgan}. 
In contrast, \method preserves the original appearance and pose of objects and generates complex and realistic shadows even without access to the full 3D model of the virtual object.



\subsection{Human perceptual study}
\label{subsec:human_evaluation}

While quantitative metrics provide a measure of image quality, recent evidence has shown they do not correlate with human perception when evaluating the realism of composited images~\cite{giroux2024towards}.
We therefore conduct two user studies to evaluate the perceived realism of the images produced by our method compared to other established approaches.

\begin{table}[!t]
\centering
\footnotesize
\begin{tabular}{lr}
\toprule
Method & Confusion (\%) \\
\midrule
\textbf{\method OR (ours)} & \textbf{45.0 $\pm$ 3.9~~~} \\
EMLight~\cite{zhan2021emlight} & 41.5 $\pm$ 3.9~~~ \\
\method IV & 35.7 $\pm$ 3.7 * \\
Garon'19~\cite{garon2019cvpr} & 31.5 $\pm$ 3.6 * \\
Everlight~\cite{dastjerdi2023everlight} & 31.4 $\pm$ 3.6 * \\
ControlCom~\cite{zhang2023controlcom} & 19.9 $\pm$ 3.1 * \\
Careaga'23~\cite{careaga2023intrinsic} & 5.0 $\pm$ 1.7 * \\
ARShadowGAN~\cite{liu2020arshadowgan} & 4.8 $\pm$ 1.7 * \\
\bottomrule
\end{tabular}
    \vspace{-5pt}
\caption{Results of our 2AFC user study indicates the perceived realism of the composites, sorted by decreasing confusion (perfect confusion is 50\%), and 95\% confidence intervals (where ``*'' indicate a statistically significant difference with \method OR).}
\label{tab:user_study}
\vspace{-7pt}
\end{table}

We first designed a two-alternative forced choice (2AFC) task where participants viewed a series of image pairs. Each pair featured a ground truth composite and a prediction from one method, both showing the same object on the same background. We asked observers, ``The same virtual object has been inserted in these two images. Click on the image that looks the most realistic'' (see supp. for instructions). To reduce bias, we randomized the left/right placement of ground truth and predicted images. Each user evaluated 20 pairs per method, totaling 160 comparisons. We randomly sampled a non-overlapping subset of the test set for each method. 
In this scenario, a confusion rate of 50\% would indicate that users find the generated composites indistinguishable from the ground truth on average. We selected the three lighting estimation methods~\cite{garon2019cvpr, zhan2021emlight, dastjerdi2023everlight} and the three image-based methods~\cite{liu2020arshadowgan, zhang2023controlcom, careaga2023intrinsic} with the highest PSNR.
A total of $N=47$ observers participated in our study.

As shown in \Cref{tab:user_study}, our method trained on OpenRooms achieves a 45\% confusion rate, indicating a strong preference for the realism of our composites. \method trained on InteriorVerse (IV) doesn't perform as well, presumably due to weaker shadows. The method with the best quantitative score from \Cref{tab:quantative_eval}, Garon'19~\cite{garon2019cvpr}, is ranked fourth with 31.5\%, corroborating recent findings that image evaluation metrics do not correlate well with human perception~\cite{giroux2024towards}. 
We achieve statistically significant better results against all methods, except with EMLight. We therefore conduct a second user study where $N=19$ participants were shown 100 pairs of images, each pair containing one result from \method and the other from EMLight, in randomized order. Users selected our method 55.4~$\pm$~2.2\% of the time, demonstrating preference for \method.

\begin{table}[t]
\centering
\footnotesize
\begin{tabular}{llcc}
\toprule
& Method & PSNR$_\uparrow$ & SSIM$_\uparrow$ \\
\midrule
\multirow{3}*{Radius}
& $\lambda=0.5$ & 32.8 & 0.973 \\
& $\lambda=1.0$ (Ours) & 31.7 & 0.970 \\
& $\lambda=1.5$ & 30.8 & 0.967 \\
\midrule
\multirow{4}*{Input} 
& w/o depth and normals & 31.8 & 0.966 \\
& w/o normal & 31.6 & 0.965 \\
& w/o depth & 32.1 & 0.969 \\
& baseline & 31.9 & 0.969 \\
\bottomrule
\end{tabular}
    \vspace{-5pt}
\caption{Ablation study on the shading radius, different inputs. The baseline and input-ablated models are trained for 220k steps.}
\label{tab:ablations}
\vspace{-7pt}
\end{table}

\subsection{Ablations} 

\begin{figure}[ht!]
    \centering
    \footnotesize
    \setlength{\tabcolsep}{1pt}
    \begin{tabular}{ccc}
        
        \begin{tikzpicture}
            
            \node[anchor=south west,inner sep=0] (image) at (0,0){\includegraphics[width=0.32\linewidth]{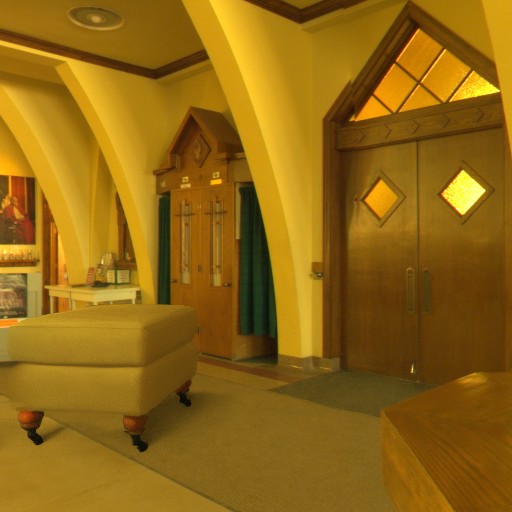}};
            \begin{scope}[x={(image.south east)},y={(image.north west)}]
                \draw[red,thick] (0.02,0.25) rectangle (0.35,0.1);
            \end{scope}
        \end{tikzpicture}
        &
        \includegraphics[width=0.32\linewidth]{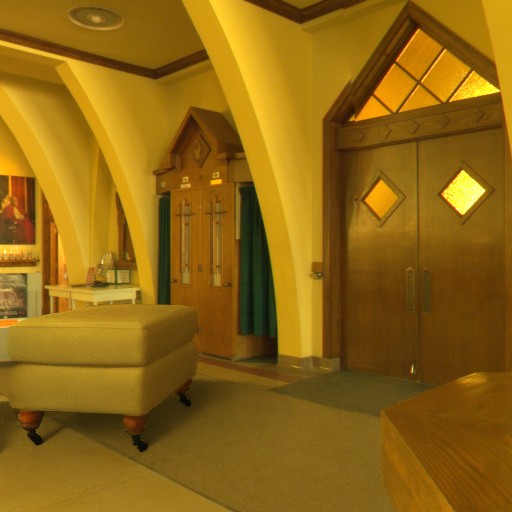}
        &
        \begin{tikzpicture}
            
            \node[anchor=south west,inner sep=0] (image) at (0,0){\includegraphics[width=0.32\linewidth]{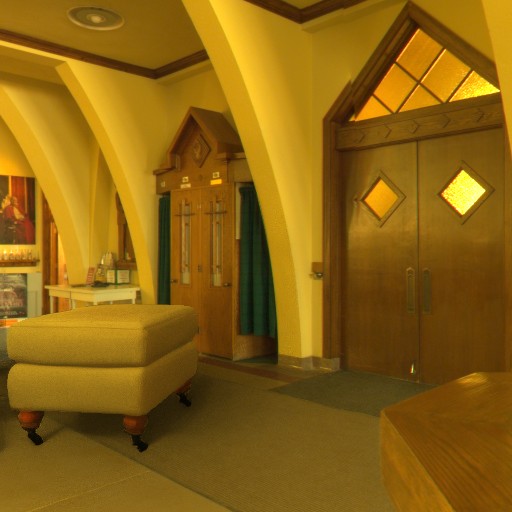}};
            \begin{scope}[x={(image.south east)},y={(image.north west)}]
                \draw[red,thick] (0.35,0.5) rectangle (0.5,0.20);
            \end{scope}
        \end{tikzpicture}
        \\        
        \includegraphics[width=0.32\linewidth]{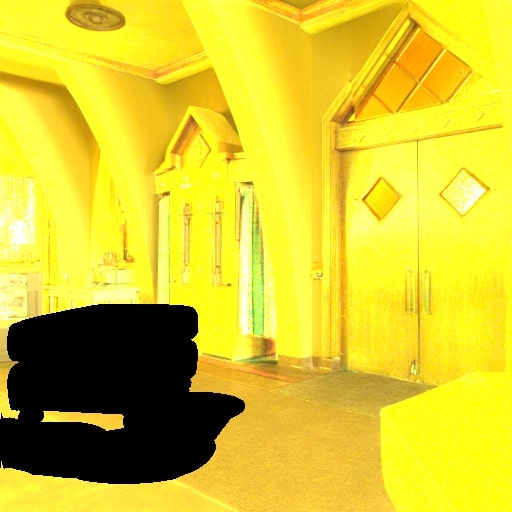} &
        \includegraphics[width=0.32\linewidth]{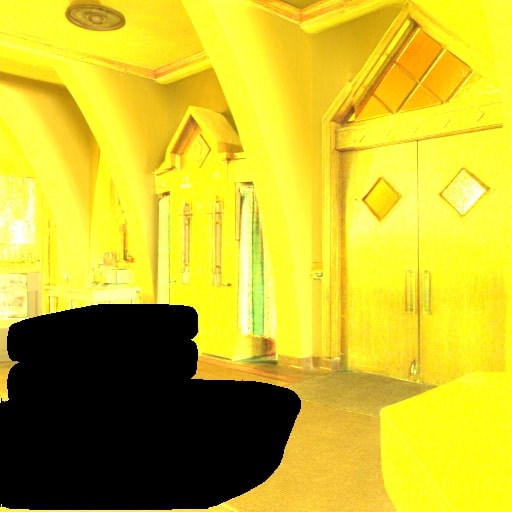} &
        \includegraphics[width=0.32\linewidth]{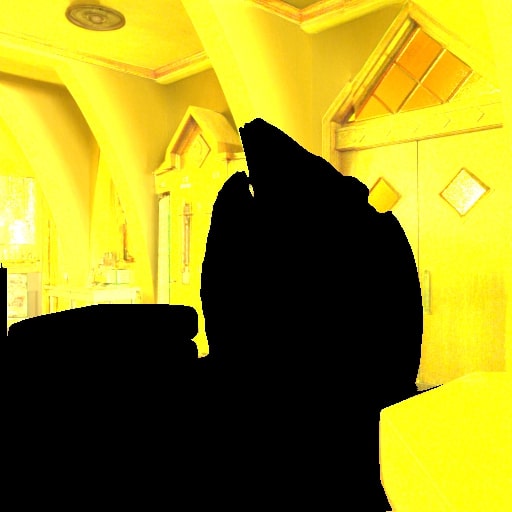} \\
        $\lambda = 0.5$ & $\lambda = 1.0$ & $\lambda = 2.0$
    \end{tabular}
    \vspace{-5pt}
    \caption{Effect of the shading mask radius $\lambda$. Generated images (top) and their associated masked shading maps (bottom) are shown. A small radius ($\lambda=0.5$) results in unrealistic shadow shapes, while a large radius ($\lambda=2.0$) produces overly large shadows and a loss of shading detail in the scene.}
    \label{fig:effect-radius}
    \vspace{-5pt}
\end{figure}

\begin{figure}
\centering
\footnotesize
\setlength{\tabcolsep}{1pt}
    \begin{tabular}{cccc}
        \includegraphics[width=0.23\linewidth]{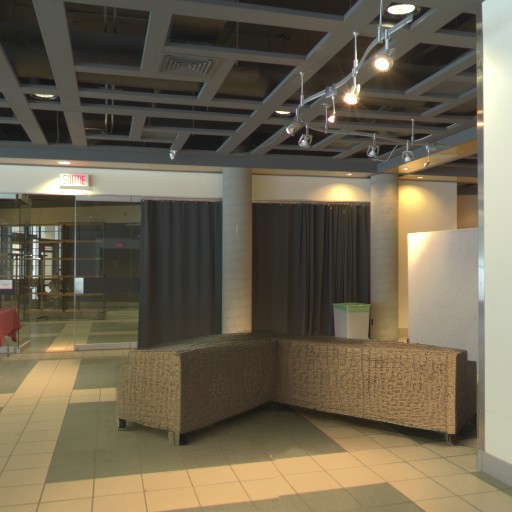} 
        \llap{\raisebox{0.9cm}{\frame{\includegraphics[height=1.2cm, trim={5cm 2cm 5cm 10cm}, clip]{figures/ablation_conditioning/df/9C4A4861-19626f89e9_04_crop_B07B4MSYPS.jpg}}}}
        &
        \includegraphics[width=0.23\linewidth]{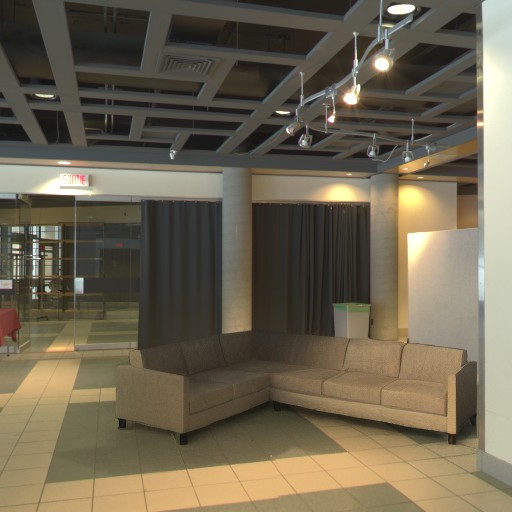} 
        \llap{\raisebox{0.9cm}{\frame{\includegraphics[height=1.2cm, trim={5cm 2cm 5cm 10cm}, clip]{figures/ablation_conditioning/df_nm/9C4A4861-19626f89e9_04_crop_B07B4MSYPS.jpg}}}}
        &
        \includegraphics[width=0.23\linewidth]{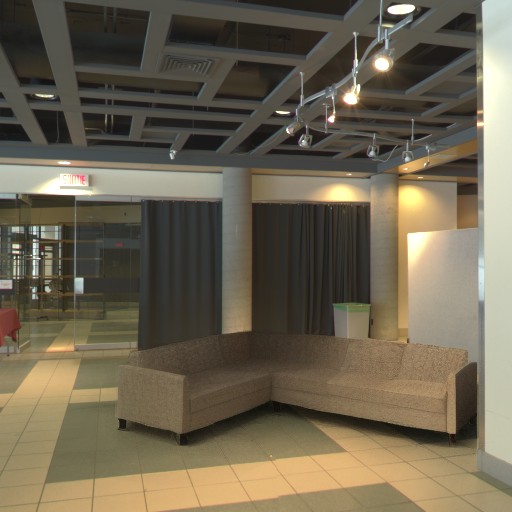} 
        \llap{\raisebox{0.9cm}{\frame{\includegraphics[height=1.2cm, trim={5cm 2cm 5cm 10cm}, clip]{figures/ablation_conditioning/df_dp/9C4A4861-19626f89e9_04_crop_B07B4MSYPS.jpg}}}}
        &
        \includegraphics[width=0.23\linewidth]{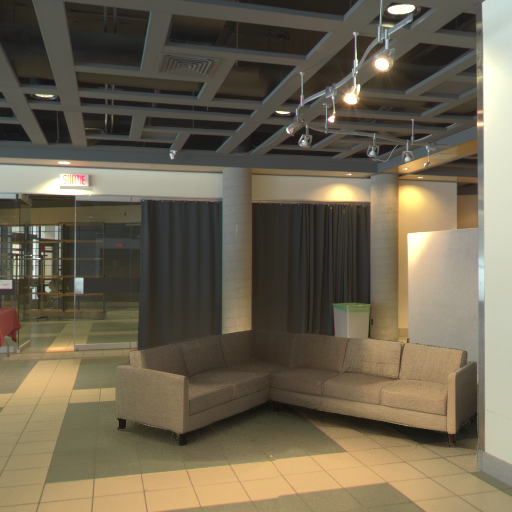} 
        \llap{\raisebox{0.9cm}{\frame{\includegraphics[height=1.2cm, trim={5cm 2cm 5cm 10cm}, clip]{figures/ablation_conditioning/df_dp_nm_2days/9C4A4861-19626f89e9_04_crop_B07B4MSYPS_pred_seed_1_comp.png}}}}
        \\
        $i_a$ & $\{i_a, i_n\}$ & $\{i_a, i_d\}$ & $\{i_a, i_n, i_d\}$
    \end{tabular}
    \vspace{-8pt}
    \caption{Ablation on intrinsic conditionings:  Training using different combinations of depth ($i_d$) and normals ($i_n$), while consistently retaining the albedo ($i_a$). Normals aid the network in generating sharp object details, whereas depth enhances shadow strength. Using both conditionings provides optimal results.}
    \label{fig:ablation-conditioning}
    \vspace{-5pt}
\end{figure}

In \Cref{tab:ablations}, we ablate shading radius (``Radius''), 
various intrinsic maps as input (``Input''). The quantitative difference due to the shading radius is confirmed visually in \Cref{fig:effect-radius}. However, metrics contradict visual observations when it comes to using different inputs. From \Cref{fig:ablation-conditioning}, not using the depth or normal maps results in a loss of realism.


\subsection{Extensions}

\begin{figure}[!th]
\centering
\scriptsize
\setlength{\tabcolsep}{1pt}
\begin{tabular}{cccc}


    \includegraphics[width=0.23\linewidth]{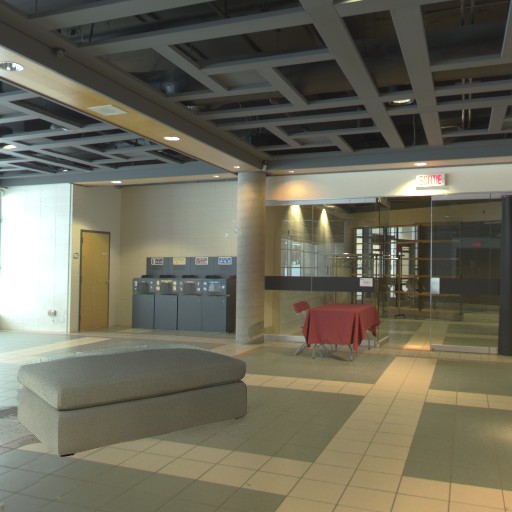} &
    \includegraphics[width=0.23\linewidth]{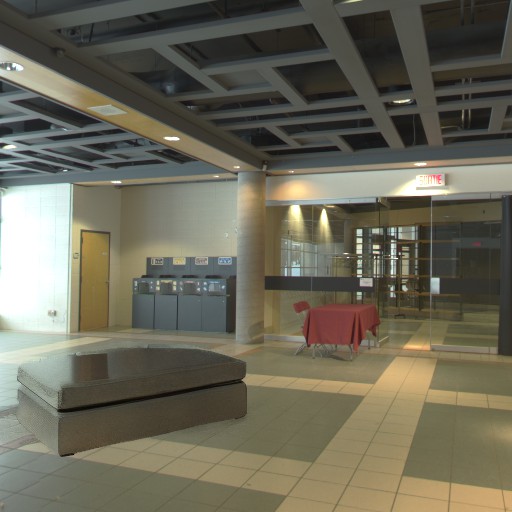} &
    \includegraphics[width=0.23\linewidth]{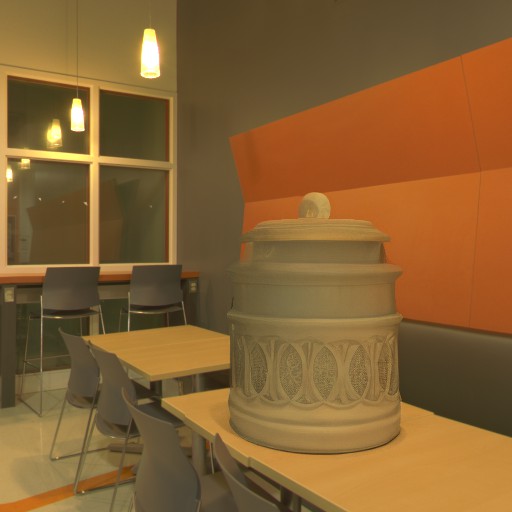} &
    \includegraphics[width=0.23\linewidth]{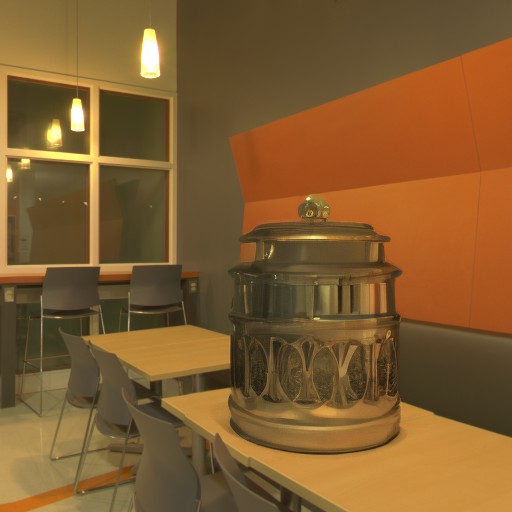} \\

Original & $RG=0,MT=1$ & Original & $RG=0,MT=1$
\end{tabular}
\vspace{-8pt}
\caption{Training \method on InteriorVerse~\cite{zhu2022learning} significantly enhances its performance with shiny objects by allowing precise control over roughness and metallic properties.}
\label{fig:advanced_materials}
\vspace{-5pt}
\end{figure}

\begin{figure}[!th]
\centering
\footnotesize
\setlength{\tabcolsep}{1pt}
\begin{tabular}{cccc}
    \includegraphics[width=0.32\linewidth]{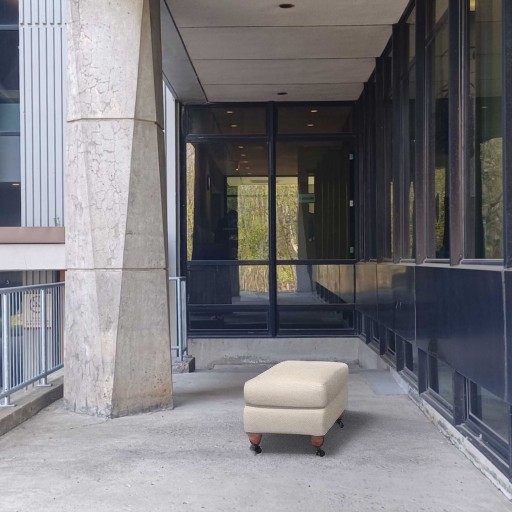} &
    \includegraphics[width=0.32\linewidth]{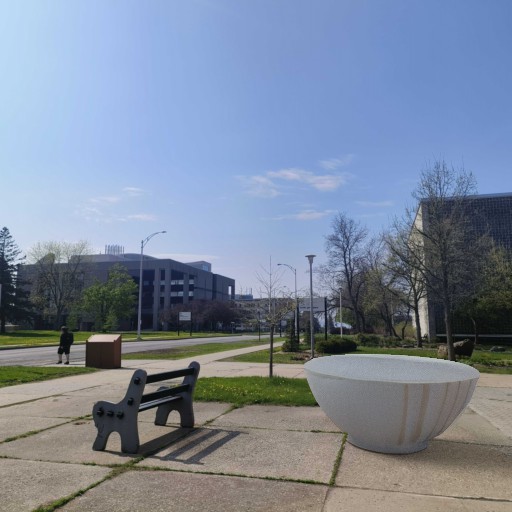} &
    \includegraphics[width=0.32\linewidth]{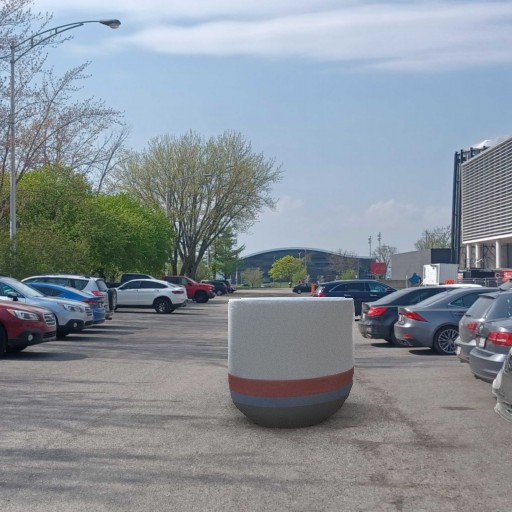} \\
\end{tabular}
\vspace{-8pt}
\caption{\method generalizes to outdoor scenes, despite being trained exclusively on indoor scenes. Note how the object shading and cast shadows seamlessly blend with the target background.}
\label{fig:outdoors}
\vspace{-7pt}
\end{figure}

\begin{figure}
    \centering
    \setlength{\tabcolsep}{1pt}
    \begin{tabular}{ccc}
       \includegraphics[width=0.32\linewidth]{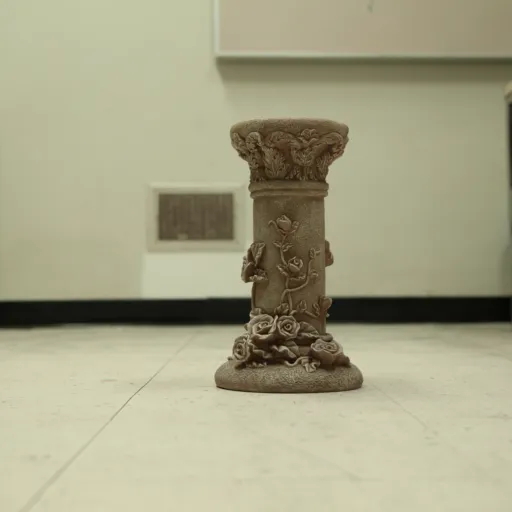} & 
       \includegraphics[width=0.32\linewidth]{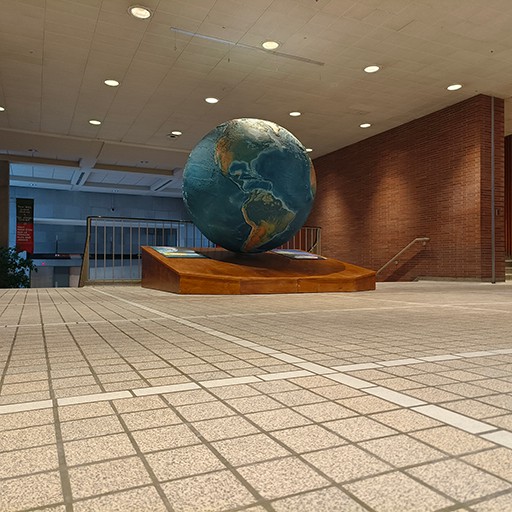} & 
       \includegraphics[width=0.32\linewidth]{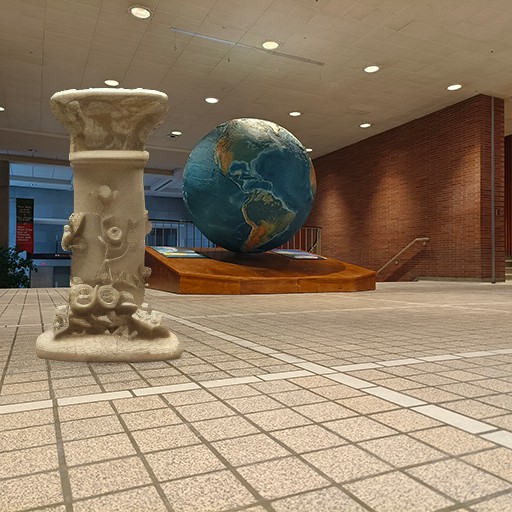} \\
       \includegraphics[width=0.32\linewidth]{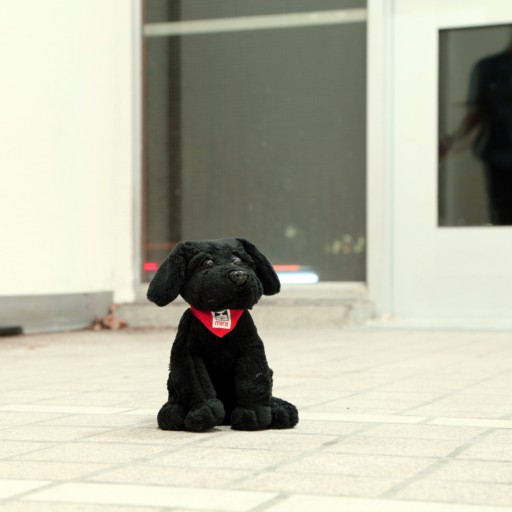} & 
       \includegraphics[width=0.32\linewidth]{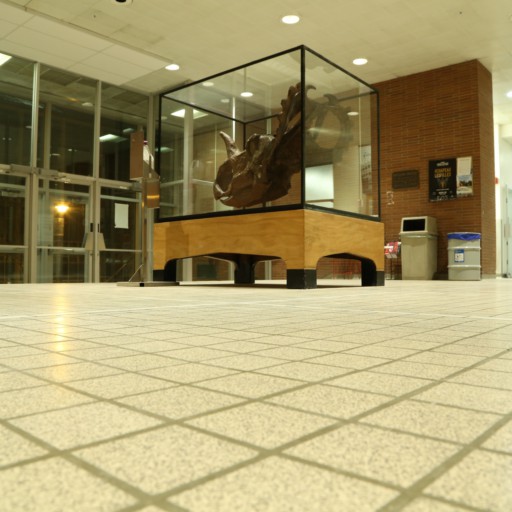} & 
       \includegraphics[width=0.32\linewidth]{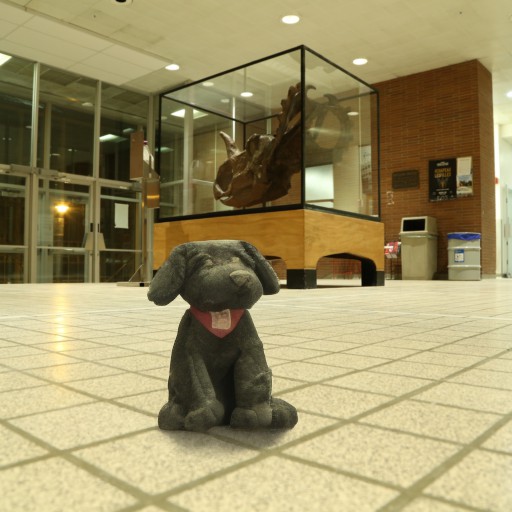} \\
    \end{tabular}
    \vspace{-7pt}
    \caption{Using \method to composite real 2D objects without access to a 3D model. Intrinsic maps for both the object image and the target background are estimated separately, then composited together and fed to our pipeline. Examples are displayed left to right: object, target background, and predicted composite.}
    \label{fig:compositing-2d}
    \vspace{-7pt}
\end{figure}

\myparagraph{Material editing.} By training on additional intrinsic maps such as roughness and metallic available in the InteriorVerse dataset \cite{zhu2022learning}, \method can also adjust the materials of the virtual object. 
In \Cref{fig:advanced_materials}, we modify the roughness (RG) and metallicity (MT) to demonstrate the effectiveness of \method in handling more advanced materials.

\myparagraph{Outdoor images.} Despite being trained only on indoor imagery, \method also generalizes to outdoor scenes and can generate realistic shadows, as demonstrated in \Cref{fig:outdoors}.

\myparagraph{2D object compositing.} \method can also be applied to 2D objects segmented from real images, where a 3D model is not available. Here, we rely on intrinsic estimators (\Cref{sec:zero-shot}) to estimate the object depth and normals. We use the RGB as the albedo to avoid detrimental noise in the image texture while keeping the rest of the pipeline unchanged. For demonstration purposes, the object was segmented and placed in the target image manually. \Cref{fig:compositing-2d} shows several such examples, showing our method can be easily extended to the case of 2D object compositing.  
\section{Discussion}

We present \method, a novel approach for creating realistic image composites with intricate lighting interactions between virtual objects and scenes. Our method achieves zero-shot compositing by training on the simpler proxy task of reconstructing an image from its intrinsic layers using readily available datasets, simplifying the training procedure. Moreover, we present a comprehensive evaluation dataset for 3D object composition in real images, where our method exhibits favorable performance compared to various light estimation and generative techniques. Through an extensive user study using the same dataset, we demonstrate that \method achieves the highest perceptual scores among recent methods. This also suggests the need for reliable quantitative metrics for lighting estimation. 

\myparagraph{Limitations.} 
\method depends on intrinsic estimators to decompose the background image. While the model shows robustness to these estimators (see supplement), even with synthetic training data, errors in their predictions can affect rendering quality. Nonetheless, \method has demonstrated impressive results when fully leveraging estimated albedo and shading for both the background and inserted objects (see \Cref{fig:compositing-2d}), suggesting that non-synthetic training data could enhance the model's robustness.

\myparagraph{Acknowledgments.}
This research was supported by NSERC grants RGPIN 2020-04799 and ALLRP 586543-23, Mitacs and Depix. Computing resources were provided by the Digital Research Alliance of Canada. The authors also thank L.-É. Messier, J. Giroux and members of the lab.


{\small
\bibliographystyle{ieee_fullname}
\bibliography{egbib}

\begin{thebibliography}{10}\itemsep=-1pt

\bibitem{andersson2020flip}
Pontus Andersson, Jim Nilsson, Tomas Akenine-M{\"o}ller, Magnus Oskarsson, Kalle {\AA}str{\"o}m, and Mark~D Fairchild.
\newblock Flip: A difference evaluator for alternating images.
\newblock {\em Proc. ACM Comput. Graph. Interact. Tech.}, 3(2):15--1, 2020.

\bibitem{bae2024dsine}
Gwangbin Bae and Andrew~J. Davison.
\newblock Rethinking inductive biases for surface normal estimation.
\newblock In {\em IEEE Conf. Comput. Vis. Pattern Recog.}, 2024.

\bibitem{barrow1978recovering}
Harry Barrow, J Tenenbaum, A Hanson, and E Riseman.
\newblock Recovering intrinsic scene characteristics.
\newblock {\em Comput. vis. syst}, 2(3-26):2, 1978.

\bibitem{bhat2023zoedepth}
Shariq~Farooq Bhat, Reiner Birkl, Diana Wofk, Peter Wonka, and Matthias M{\"u}ller.
\newblock Zoedepth: Zero-shot transfer by combining relative and metric depth.
\newblock {\em arXiv preprint arXiv:2302.12288}, 2023.

\bibitem{bhattad2022cut}
Anand Bhattad and David~A Forsyth.
\newblock Cut-and-paste object insertion by enabling deep image prior for reshading.
\newblock In {\em Int. Conf. 3D Vis.}, 2022.

\bibitem{bhattad2023stylegan}
Anand Bhattad, Daniel McKee, Derek Hoiem, and David Forsyth.
\newblock Stylegan knows normal, depth, albedo, and more.
\newblock {\em Advances in Neural Information Processing Systems}, 36, 2023.

\bibitem{bhattad2023make}
Anand Bhattad, Viraj Shah, Derek Hoiem, and David~A Forsyth.
\newblock Make it so: Steering stylegan for any image inversion and editing.
\newblock {\em arXiv preprint arXiv:2304.14403}, 2023.

\bibitem{bhattad2024stylitgan}
Anand Bhattad, James Soole, and DA Forsyth.
\newblock Stylitgan: Image-based relighting via latent control.
\newblock In {\em IEEE Conf. Comput. Vis. Pattern Recog.}, 2024.

\bibitem{burt1983multiresolution}
Peter~J Burt and Edward~H Adelson.
\newblock A multiresolution spline with application to image mosaics.
\newblock {\em ACM Trans. Graph.}, 2(4):217--236, 1983.

\bibitem{careaga2023ordinal}
Chris Careaga and Ya{\u{g}}{\i}z Aksoy.
\newblock Intrinsic image decomposition via ordinal shading.
\newblock {\em ACM Trans. Graph.}, 43(1):1--24, 2023.

\bibitem{careaga2023intrinsic}
Chris Careaga, S~Mahdi~H Miangoleh, and Ya{\u{g}}{\i}z Aksoy.
\newblock Intrinsic harmonization for illumination-aware compositing.
\newblock {\em ACM Trans. Graph.}, 2023.

\bibitem{chen2023anydoor}
Xi Chen, Lianghua Huang, Yu Liu, Yujun Shen, Deli Zhao, and Hengshuang Zhao.
\newblock Anydoor: Zero-shot object-level image customization.
\newblock {\em arXiv preprint arXiv:2307.09481}, 2023.

\bibitem{choi2023mair}
JunYong Choi, SeokYeong Lee, Haesol Park, Seung-Won Jung, Ig-Jae Kim, and Junghyun Cho.
\newblock Mair: multi-view attention inverse rendering with 3d spatially-varying lighting estimation.
\newblock In {\em IEEE Conf. Comput. Vis. Pattern Recog.}, 2023.

\bibitem{chuang2003shadow}
Yung-Yu Chuang, Dan~B Goldman, Brian Curless, David~H Salesin, and Richard Szeliski.
\newblock Shadow matting and compositing.
\newblock In {\em ACM SIGGRAPH Conf.}, 2003.

\bibitem{collins2022abo}
Jasmine Collins, Shubham Goel, Kenan Deng, Achleshwar Luthra, Leon Xu, Erhan Gundogdu, Xi Zhang, Tomas~F Yago~Vicente, Thomas Dideriksen, Himanshu Arora, Matthieu Guillaumin, and Jitendra Malik.
\newblock {ABO}: Dataset and benchmarks for real-world 3d object understanding.
\newblock In {\em IEEE Conf. Comput. Vis. Pattern Recog.}, 2022.

\bibitem{blender}
Blender~Online Community.
\newblock {\em Blender - a 3D modelling and rendering package}.
\newblock Blender Foundation, Stichting Blender Foundation, Amsterdam, 2018.

\bibitem{dastjerdi2023everlight}
Mohammad Reza~Karimi Dastjerdi, Jonathan Eisenmann, Yannick Hold-Geoffroy, and Jean-Fran{\c{c}}ois Lalonde.
\newblock Everlight: Indoor-outdoor editable {HDR} lighting estimation.
\newblock In {\em Int. Conf. Comput. Vis.}, 2023.

\bibitem{debevec1998rendering}
Paul Debevec.
\newblock Rendering synthetic objects into real scenes: Bridging traditional and image-based graphics with global illumination and high dynamic range photography.
\newblock In {\em Conf. Comp. Graph. Int. Tech.}, ACM SIGGRAPH Conf., pages 189--198, 1998.

\bibitem{du2023generative}
Xiaodan Du, Nicholas Kolkin, Greg Shakhnarovich, and Anand Bhattad.
\newblock Generative models: What do they know? do they know things? let's find out!
\newblock {\em arXiv preprint arXiv:2311.17137}, 2023.

\bibitem{garces2022survey}
Elena Garces, Carlos Rodriguez-Pardo, Dan Casas, and Jorge Lopez-Moreno.
\newblock A survey on intrinsic images: Delving deep into lambert and beyond.
\newblock {\em International Journal of Computer Vision}, 130(3):836--868, 2022.

\bibitem{gardner2019iccv}
Marc-Andr{\'e} Gardner, Yannick Hold-Geoffroy, Kalyan Sunkavalli, Christian Gagn{\'e}, and Jean-Fran{\c{c}}ois Lalonde.
\newblock Deep parametric indoor lighting estimation.
\newblock In {\em Int. Conf. Comput. Vis.}, 2019.

\bibitem{gardner2017sigasia}
Marc-Andr\'{e} Gardner, Kalyan Sunkavalli, Ersin Yumer, Xiaohui Shen, Emiliano Gambaretto, Christian Gagn\'{e}, and Jean-Fran\c{c}ois Lalonde.
\newblock Learning to predict indoor illumination from a single image.
\newblock {\em ACM Trans. Graph.}, 9(4), 2017.

\bibitem{garon2019cvpr}
Mathieu Garon, Kalyan Sunkavalli, Sunil Hadap, Nathan Carr, and Jean-Fran{\c{c}}ois Lalonde.
\newblock Fast spatially-varying indoor lighting estimation.
\newblock In {\em IEEE Conf. Comput. Vis. Pattern Recog.}, 2019.

\bibitem{ge20243d}
Yunhao Ge, Hong-Xing Yu, Cheng Zhao, Yuliang Guo, Xinyu Huang, Liu Ren, Laurent Itti, and Jiajun Wu.
\newblock 3d copy-paste: Physically plausible object insertion for monocular 3d detection.
\newblock In {\em Adv. Neural Inform. Process. Syst.}, 2024.

\bibitem{geng2023tree}
Chen Geng, Hong-Xing Yu, Sharon Zhang, Maneesh Agrawala, and Jiajun Wu.
\newblock Tree-structured shading decomposition.
\newblock In {\em Int. Conf. Comput. Vis.}, pages 488--498, 2023.

\bibitem{ghazanfarilipsim}
Sara Ghazanfari, Alexandre Araujo, Prashanth Krishnamurthy, Farshad Khorrami, and Siddharth Garg.
\newblock Lipsim: A provably robust perceptual similarity metric.
\newblock In {\em Int. Conf. Learn. Represent.}, 2024.

\bibitem{giroux2024towards}
Justine Giroux, Mohammad Reza~Karimi Dastjerdi, Yannick Hold-Geoffroy, Javier Vazquez-Corral, and Jean-Fran\c{c}ois Lalonde.
\newblock Towards a perceptual evaluation framework for lighting estimation.
\newblock In {\em IEEE Conf. Comput. Vis. Pattern Recog.}, 2024.

\bibitem{goodfellow2014explaining}
Ian~J Goodfellow, Jonathon Shlens, and Christian Szegedy.
\newblock Explaining and harnessing adversarial examples.
\newblock {\em arXiv preprint arXiv:1412.6572}, 2014.

\bibitem{guo2021intrinsic}
Zonghui Guo, Haiyong Zheng, Yufeng Jiang, Zhaorui Gu, and Bing Zheng.
\newblock Intrinsic image harmonization.
\newblock In {\em IEEE Conf. Comput. Vis. Pattern Recog.}, 2021.

\bibitem{hold2017cvpr}
Yannick Hold-Geoffroy, Kalyan Sunkavalli, Sunil Hadap, Emiliano Gambaretto, and Jean-Fran{\c{c}}ois Lalonde.
\newblock Deep outdoor illumination estimation.
\newblock In {\em IEEE Conf. Comput. Vis. Pattern Recog.}, 2017.

\bibitem{hu2021lora}
Edward~J Hu, Phillip Wallis, Zeyuan Allen-Zhu, Yuanzhi Li, Shean Wang, Lu Wang, Weizhu Chen, et~al.
\newblock Lora: Low-rank adaptation of large language models.
\newblock In {\em Int. Conf. Learn. Represent.}, 2021.

\bibitem{jin2023perspective}
Linyi Jin, Jianming Zhang, Yannick Hold-Geoffroy, Oliver Wang, Kevin Blackburn-Matzen, Matthew Sticha, and David~F Fouhey.
\newblock Perspective fields for single image camera calibration.
\newblock In {\em IEEE Conf. Comput. Vis. Pattern Recog.}, 2023.

\bibitem{karsch2011rendering}
Kevin Karsch, Varsha Hedau, David Forsyth, and Derek Hoiem.
\newblock Rendering synthetic objects into legacy photographs.
\newblock {\em ACM Trans. Graph.}, 30(6):1--12, 2011.

\bibitem{karsch2014automatic}
Kevin Karsch, Kalyan Sunkavalli, Sunil Hadap, Nathan Carr, Hailin Jin, Rafael Fonte, Michael Sittig, and David Forsyth.
\newblock Automatic scene inference for 3d object compositing.
\newblock {\em ACM Trans. Graph.}, 33(3):1--15, 2014.

\bibitem{kettunen2019lpips}
Markus Kettunen, Erik H{\"a}rk{\"o}nen, and Jaakko Lehtinen.
\newblock E-lpips: robust perceptual image similarity via random transformation ensembles.
\newblock {\em arXiv preprint arXiv:1906.03973}, 2019.

\bibitem{kocsis2024lightit}
Peter Kocsis, Julien Philip, Kalyan Sunkavalli, Matthias Nie{\ss}ner, and Yannick Hold-Geoffroy.
\newblock Lightit: Illumination modeling and control for diffusion models.
\newblock In {\em IEEE Conf. Comput. Vis. Pattern Recog.}, 2024.

\bibitem{kocsis2024intrinsic}
Peter Kocsis, Vincent Sitzmann, and Matthias Nie{\ss}ner.
\newblock Intrinsic image diffusion for indoor single-view material estimation.
\newblock In {\em IEEE Conf. Comput. Vis. Pattern Recog.}, pages 5198--5208, 2024.

\bibitem{kocsis2023iid}
Peter Kocsis, Vincent Sitzmann, and Matthias Nie{\ss}ner.
\newblock Intrinsic image diffusion for single-view material estimation.
\newblock In {\em IEEE Conf. Comput. Vis. Pattern Recog.}, 2024.

\bibitem{kovacs2017shading}
Balazs Kovacs, Sean Bell, Noah Snavely, and Kavita Bala.
\newblock Shading annotations in the wild.
\newblock In {\em IEEE Conf. Comput. Vis. Pattern Recog.}, pages 6998--7007, 2017.

\bibitem{lalonde2007using}
Jean-Francois Lalonde and Alexei~A Efros.
\newblock Using color compatibility for assessing image realism.
\newblock In {\em Int. Conf. Comput. Vis.}, 2007.

\bibitem{lalonde2007photo}
Jean-Fran{\c{c}}ois Lalonde, Derek Hoiem, Alexei~A Efros, Carsten Rother, John Winn, and Antonio Criminisi.
\newblock Photo clip art.
\newblock {\em ACM Trans. Graph.}, 26(3), 2007.

\bibitem{li2020cvpr}
Zhengqin Li, Mohammad Shafiei, Ravi Ramamoorthi, Kalyan Sunkavalli, and Manmohan Chandraker.
\newblock Inverse rendering for complex indoor scenes: Shape, spatially-varying lighting and svbrdf from a single image.
\newblock In {\em IEEE Conf. Comput. Vis. Pattern Recog.}, 2020.

\bibitem{li2022physically}
Zhengqin Li, Jia Shi, Sai Bi, Rui Zhu, Kalyan Sunkavalli, Milo{\v{s}} Ha{\v{s}}an, Zexiang Xu, Ravi Ramamoorthi, and Manmohan Chandraker.
\newblock Physically-based editing of indoor scene lighting from a single image.
\newblock In {\em Eur. Conf. Comput. Vis.}, 2022.

\bibitem{li2020openrooms}
Zhengqin Li, Ting-Wei Yu, Shen Sang, Sarah Wang, Meng Song, Yuhan Liu, Yu-Ying Yeh, Rui Zhu, Nitesh Gundavarapu, Jia Shi, et~al.
\newblock Openrooms: An end-to-end open framework for photorealistic indoor scene datasets.
\newblock In {\em IEEE Conf. Comput. Vis. Pattern Recog.}, 2021.

\bibitem{lin2024common}
Shanchuan Lin, Bingchen Liu, Jiashi Li, and Xiao Yang.
\newblock Common diffusion noise schedules and sample steps are flawed.
\newblock In {\em IEEE/CVF Winter Conf. App. Comp. Vis.}, 2024.

\bibitem{liu2020arshadowgan}
Daquan Liu, Chengjiang Long, Hongpan Zhang, Hanning Yu, Xinzhi Dong, and Chunxia Xiao.
\newblock Arshadowgan: Shadow generative adversarial network for augmented reality in single light scenes.
\newblock In {\em IEEE Conf. Comput. Vis. Pattern Recog.}, 2020.

\bibitem{luo2024intrinsicdiffusion}
Jundan Luo, Duygu Ceylan, Jae~Shin Yoon, Nanxuan Zhao, Julien Philip, Anna Fr{\"u}hst{\"u}ck, Wenbin Li, Christian Richardt, and Tuanfeng Wang.
\newblock Intrinsicdiffusion: Joint intrinsic layers from latent diffusion models.
\newblock In {\em ACM SIGGRAPH Conf.}, pages 1--11, 2024.

\bibitem{michel2024object}
Oscar Michel, Anand Bhattad, Eli VanderBilt, Ranjay Krishna, Aniruddha Kembhavi, and Tanmay Gupta.
\newblock Object 3dit: Language-guided 3d-aware image editing.
\newblock In {\em Adv. Neural Inform. Process. Syst.}, 2024.

\bibitem{perez2003poisson}
Patrick P{\'e}rez, Michel Gangnet, and Andrew Blake.
\newblock Poisson image editing.
\newblock {\em ACM Trans. Graph.}, 22(3), 2003.

\bibitem{philip2019multi}
Julien Philip, Micha{\"e}l Gharbi, Tinghui Zhou, Alexei~A Efros, and George Drettakis.
\newblock Multi-view relighting using a geometry-aware network.
\newblock {\em ACM Trans. Graph.}, 38(4):78--1, 2019.

\bibitem{phongthawee2024diffusionlight}
Pakkapon Phongthawee, Worameth Chinchuthakun, Nontaphat Sinsunthithet, Varun Jampani, Amit Raj, Pramook Khungurn, and Supasorn Suwajanakorn.
\newblock Diffusionlight: Light probes for free by painting a chrome ball.
\newblock In {\em IEEE Conf. Comput. Vis. Pattern Recog.}, 2024.

\bibitem{pitie2005n}
Francois Pitie, Anil~C Kokaram, and Rozenn Dahyot.
\newblock N-dimensional probability density function transfer and its application to color transfer.
\newblock In {\em Int. Conf. Comput. Vis.}, 2005.

\bibitem{poirier2024diffusion}
Yohan Poirier-Ginter, Alban Gauthier, Julien Phillip, Jean-Fran{\c{c}}ois Lalonde, and George Drettakis.
\newblock A diffusion approach to radiance field relighting using multi-illumination synthesis.
\newblock {\em Comput. Graph. Forum}, 43(4), 2024.

\bibitem{ramanagopal2024theory}
Mani Ramanagopal, Sriram Narayanan, Aswin~C Sankaranarayanan, and Srinivasa~G Narasimhan.
\newblock A theory of joint light and heat transport for lambertian scenes.
\newblock In {\em IEEE Conf. Comput. Vis. Pattern Recog.}, pages 11924--11933, 2024.

\bibitem{reinhard2001color}
Erik Reinhard, Michael Adhikhmin, Bruce Gooch, and Peter Shirley.
\newblock Color transfer between images.
\newblock {\em IEEE Comp. Graph. Appl.}, 21(5):34--41, 2001.

\bibitem{rey2022360monodepth}
Manuel Rey-Area, Mingze Yuan, and Christian Richardt.
\newblock 360monodepth: High-resolution $360^\circ$ monocular depth estimation.
\newblock In {\em IEEE Conf. Comput. Vis. Pattern Recog.}, 2022.

\bibitem{roberts2021hypersim}
Mike Roberts, Jason Ramapuram, Anurag Ranjan, Atulit Kumar, Miguel~Angel Bautista, Nathan Paczan, Russ Webb, and Joshua~M. Susskind.
\newblock Hypersim: A photorealistic synthetic dataset for holistic indoor scene understanding.
\newblock In {\em Int. Conf. Comput. Vis.}, 2021.

\bibitem{rombach2022high}
Robin Rombach, Andreas Blattmann, Dominik Lorenz, Patrick Esser, and Bj{\"o}rn Ommer.
\newblock High-resolution image synthesis with latent diffusion models.
\newblock In {\em IEEE Conf. Comput. Vis. Pattern Recog.}, 2022.

\bibitem{rother2004grabcut}
Carsten Rother, Vladimir Kolmogorov, and Andrew Blake.
\newblock Grabcut interactive foreground extraction using iterated graph cuts.
\newblock {\em ACM Trans. Graph.}, 23(3):309--314, 2004.

\bibitem{sharma2024alchemist}
Prafull Sharma, Varun Jampani, Yuanzhen Li, Xuhui Jia, Dmitry Lagun, Fredo Durand, Bill Freeman, and Mark Matthews.
\newblock Alchemist: Parametric control of material properties with diffusion models.
\newblock In {\em IEEE Conf. Comput. Vis. Pattern Recog.}, 2024.

\bibitem{sheng2022controllable}
Yichen Sheng, Yifan Liu, Jianming Zhang, Wei Yin, A~Cengiz Oztireli, He Zhang, Zhe Lin, Eli Shechtman, and Bedrich Benes.
\newblock Controllable shadow generation using pixel height maps.
\newblock In {\em Eur. Conf. Comput. Vis.}, 2022.

\bibitem{sheng2021ssn}
Yichen Sheng, Jianming Zhang, and Bedrich Benes.
\newblock Ssn: Soft shadow network for image compositing.
\newblock In {\em IEEE Conf. Comput. Vis. Pattern Recog.}, 2021.

\bibitem{sheng2023pixht}
Yichen Sheng, Jianming Zhang, Julien Philip, Yannick Hold-Geoffroy, Xin Sun, He Zhang, Lu Ling, and Bedrich Benes.
\newblock Pixht-lab: Pixel height based light effect generation for image compositing.
\newblock In {\em IEEE Conf. Comput. Vis. Pattern Recog.}, 2023.

\bibitem{song2022objectstitch}
Yizhi Song, Zhifei Zhang, Zhe Lin, Scott Cohen, Brian Price, Jianming Zhang, Soo~Ye Kim, and Daniel Aliaga.
\newblock Objectstitch: Generative object compositing.
\newblock In {\em IEEE Conf. Comput. Vis. Pattern Recog.}, 2022.

\bibitem{srinivasan2020lighthouse}
Pratul~P Srinivasan, Ben Mildenhall, Matthew Tancik, Jonathan~T Barron, Richard Tucker, and Noah Snavely.
\newblock Lighthouse: Predicting lighting volumes for spatially-coherent illumination.
\newblock In {\em IEEE Conf. Comput. Vis. Pattern Recog.}, 2020.

\bibitem{sunkavalli2010multi}
Kalyan Sunkavalli, Micah~K Johnson, Wojciech Matusik, and Hanspeter Pfister.
\newblock Multi-scale image harmonization.
\newblock {\em ACM Trans. Graph.}, 29(4), 2010.

\bibitem{tao2013error}
Michael~W Tao, Micah~K Johnson, and Sylvain Paris.
\newblock Error-tolerant image compositing.
\newblock {\em Int. J. Comput. Vis.}, 103:178--189, 2013.

\bibitem{tsai2017deep}
Yi-Hsuan Tsai, Xiaohui Shen, Zhe Lin, Kalyan Sunkavalli, Xin Lu, and Ming-Hsuan Yang.
\newblock Deep image harmonization.
\newblock In {\em IEEE Conf. Comput. Vis. Pattern Recog.}, 2017.

\bibitem{valencca2023shadow}
Lucas Valen{\c{c}}a, Jinsong Zhang, Micha{\"e}l Gharbi, Yannick Hold-Geoffroy, and Jean-Fran{\c{c}}ois Lalonde.
\newblock Shadow harmonization for realistic compositing.
\newblock In {\em ACM SIGGRAPH Asia Conf.}, 2023.

\bibitem{wang2022stylelight}
Guangcong Wang, Yinuo Yang, Chen~Change Loy, and Ziwei Liu.
\newblock Stylelight: {HDR} panorama generation for lighting estimation and editing.
\newblock In {\em Eur. Conf. Comput. Vis.}, 2022.

\bibitem{wang2004ssim}
Zhou Wang, Alan~C. Bovik, Hamid~R. Sheikh, and Eero~P. Simoncelli.
\newblock Image quality assessment: from error visibility to structural similarity.
\newblock {\em IEEE Trans. Image Process.}, 13(4):600--612, 2004.

\bibitem{weber2022editable}
Henrique Weber, Mathieu Garon, and Jean-Fran{\c{c}}ois Lalonde.
\newblock Editable indoor lighting estimation.
\newblock In {\em Eur. Conf. Comput. Vis.}, 2022.

\bibitem{wu2023measured}
Jiaye Wu, Sanjoy Chowdhury, Hariharmano Shanmugaraja, David Jacobs, and Soumyadip Sengupta.
\newblock Measured albedo in the wild: Filling the gap in intrinsics evaluation.
\newblock In {\em 2023 IEEE International Conference on Computational Photography (ICCP)}, pages 1--12. IEEE, 2023.

\bibitem{xu2024goodseedmakesgood}
Katherine Xu, Lingzhi Zhang, and Jianbo Shi.
\newblock Good seed makes a good crop: Discovering secret seeds in text-to-image diffusion models, 2024.

\bibitem{yang2023paint}
Binxin Yang, Shuyang Gu, Bo Zhang, Ting Zhang, Xuejin Chen, Xiaoyan Sun, Dong Chen, and Fang Wen.
\newblock Paint by example: Exemplar-based image editing with diffusion models.
\newblock In {\em IEEE Conf. Comput. Vis. Pattern Recog.}, 2023.

\bibitem{ye2024stablenormal}
Chongjie Ye, Lingteng Qiu, Xiaodong Gu, Qi Zuo, Yushuang Wu, Zilong Dong, Liefeng Bo, Yuliang Xiu, and Xiaoguang Han.
\newblock Stablenormal: Reducing diffusion variance for stable and sharp normal.
\newblock {\em arXiv preprint arXiv:2406.16864}, 2024.

\bibitem{ye2023ip}
Hu Ye, Jun Zhang, Sibo Liu, Xiao Han, and Wei Yang.
\newblock Ip-adapter: Text compatible image prompt adapter for text-to-image diffusion models.
\newblock {\em arXiv preprint arXiv:2308.06721}, 2023.

\bibitem{ye2023intrinsicnerf}
Weicai Ye, Shuo Chen, Chong Bao, Hujun Bao, Marc Pollefeys, Zhaopeng Cui, and Guofeng Zhang.
\newblock Intrinsicnerf: Learning intrinsic neural radiance fields for editable novel view synthesis.
\newblock In {\em Int. Conf. Comput. Vis.}, pages 339--351, 2023.

\bibitem{yoshida2023light}
Yusaku Yoshida, Ryo Kawahara, and Takahiro Okabe.
\newblock Light source separation and intrinsic image decomposition under ac illumination.
\newblock In {\em IEEE Conf. Comput. Vis. Pattern Recog.}, pages 5735--5743, 2023.

\bibitem{yuan2023customnet}
Ziyang Yuan, Mingdeng Cao, Xintao Wang, Zhongang Qi, Chun Yuan, and Ying Shan.
\newblock Customnet: Zero-shot object customization with variable-viewpoints in text-to-image diffusion models.
\newblock {\em arXiv preprint arXiv:2310.19784}, 2023.

\bibitem{zamir2019vision}
Syed~Waqas Zamir, Javier Vazquez-Corral, and Marcelo Bertalmio.
\newblock Vision models for wide color gamut imaging in cinema.
\newblock {\em IEEE Trans. Pattern Anal. Mach. Intell.}, 43(5):1777--1790, 2019.

\bibitem{zeng2024dilightnet}
Chong Zeng, Yue Dong, Pieter Peers, Youkang Kong, Hongzhi Wu, and Xin Tong.
\newblock {DiLightNet}: Fine-grained lighting control for diffusion-based image generation.
\newblock In {\em ACM SIGGRAPH Conf.}, 2024.

\bibitem{zeng2024rgb}
Zheng Zeng, Valentin Deschaintre, Iliyan Georgiev, Yannick Hold-Geoffroy, Yiwei Hu, Fujun Luan, Ling-Qi Yan, and Milo{\v{s}} Ha{\v{s}}an.
\newblock {RGB} $\leftrightarrow$ {X}: Image decomposition and synthesis using material-and lighting-aware diffusion models.
\newblock In {\em ACM SIGGRAPH Conf.}, 2024.

\bibitem{zhan2021emlight}
Fangneng Zhan, Changgong Zhang, Yingchen Yu, Yuan Chang, Shijian Lu, Feiying Ma, and Xuansong Xie.
\newblock {EMLight: Lighting Estimation via Spherical Distribution Approximation}.
\newblock In {\em AAAI}, 2021.

\bibitem{zhang2023controlcom}
Bo Zhang, Yuxuan Duan, Jun Lan, Yan Hong, Huijia Zhu, Weiqiang Wang, and Li Niu.
\newblock Controlcom: Controllable image composition using diffusion model.
\newblock {\em arXiv preprint arXiv:2308.10040}, 2023.

\bibitem{zhang2023adding}
Lvmin Zhang, Anyi Rao, and Maneesh Agrawala.
\newblock Adding conditional control to text-to-image diffusion models.
\newblock In {\em Int. Conf. Comput. Vis.}, 2023.

\bibitem{zhang2018lpips}
Richard Zhang, Phillip Isola, Alexei~A Efros, Eli Shechtman, and Oliver Wang.
\newblock The unreasonable effectiveness of deep features as a perceptual metric.
\newblock In {\em IEEE Conf. Comput. Vis. Pattern Recog.}, 2018.

\bibitem{zhang2023paste}
Xin Zhang, Jiaxian Guo, Paul Yoo, Yutaka Matsuo, and Yusuke Iwasawa.
\newblock Paste, inpaint and harmonize via denoising: Subject-driven image editing with pre-trained diffusion model.
\newblock {\em arXiv preprint arXiv:2306.07596}, 2023.

\bibitem{zhao2024illuminerf}
Xiaoming Zhao, Pratul~P. Srinivasan, Dor Verbin, Keunhong Park, Ricardo~Martin Brualla, and Philipp Henzler.
\newblock {IllumiNeRF}: {3D} relighting without inverse rendering.
\newblock {\em ArXiv}, 2024.

\bibitem{zhou2015learning}
Tinghui Zhou, Philipp Krahenbuhl, and Alexei~A Efros.
\newblock Learning data-driven reflectance priors for intrinsic image decomposition.
\newblock In {\em Int. Conf. Comput. Vis.}, pages 3469--3477, 2015.

\bibitem{zhu2022learning}
Jingsen Zhu, Fujun Luan, Yuchi Huo, Zihao Lin, Zhihua Zhong, Dianbing Xi, Rui Wang, Hujun Bao, Jiaxiang Zheng, and Rui Tang.
\newblock Learning-based inverse rendering of complex indoor scenes with differentiable monte carlo raytracing.
\newblock In {\em ACM SIGGRAPH Asia Conf.} ACM, 2022.

\bibitem{zhu2022irisformer}
Rui Zhu, Zhengqin Li, Janarbek Matai, Fatih Porikli, and Manmohan Chandraker.
\newblock Irisformer: Dense vision transformers for single-image inverse rendering in indoor scenes.
\newblock In {\em IEEE Conf. Comput. Vis. Pattern Recog.}, 2022.

\bibitem{zhu2023designing}
Zixin Zhu, Xuelu Feng, Dongdong Chen, Jianmin Bao, Le Wang, Yinpeng Chen, Lu Yuan, and Gang Hua.
\newblock Designing a better asymmetric vqgan for stablediffusion.
\newblock {\em arXiv preprint arXiv:2306.04632}, 2023.

\bibitem{zoran2015learning}
Daniel Zoran, Phillip Isola, Dilip Krishnan, and William~T Freeman.
\newblock Learning ordinal relationships for mid-level vision.
\newblock In {\em Int. Conf. Comput. Vis.}, pages 388--396, 2015.

\end{thebibliography}
}

\end{document}